\documentclass{article}

\usepackage{microtype}
\usepackage{graphicx}
\usepackage{booktabs} 
\usepackage{multirow}
\usepackage{amsmath,amssymb}
\usepackage{booktabs}
\usepackage{caption,subcaption}
\usepackage{xcolor}
\usepackage{enumitem}
\setitemize{itemsep=10pt,topsep=0pt,parsep=0pt,partopsep=0pt}
\usepackage{hyperref}
\graphicspath{{figures/}}

\usepackage{amsmath}
\usepackage{amsfonts}
\usepackage{amssymb}
\usepackage{wrapfig}
\usepackage{subcaption}
\usepackage{multirow}
 \usepackage{mathtools} 

\usepackage{verbatim}

\usepackage{anyfontsize}

\usepackage{color}
\usepackage{tikz}
\usetikzlibrary{arrows,shapes,snakes,automata,backgrounds,fit,petri}
\usepackage{adjustbox}

\newcommand{\RN}[1]{%
	\textup{\lowercase\expandafter{\it \romannumeral#1}}%
}

\usepackage{booktabs}

\usepackage{tcolorbox}
\usepackage{colortbl}

\makeatletter
\newcommand{\distas}[1]{\mathbin{\overset{#1}{\kern\z@\sim}}}%

\usepackage{enumitem}

\usepackage[lined,boxed,commentsnumbered,ruled,linesnumbered]{algorithm2e}

\newcommand{\ie}[0]{\emph{i.e., }}

\newcommand{\eg}[0]{\emph{e.g., }}

\newcommand{\beq}{\vspace{0mm}\begin{equation}}
\newcommand{\eeq}{\vspace{0mm}\end{equation}}
\newcommand{\beqs}{\vspace{0mm}\begin{eqnarray}}
\newcommand{\eeqs}{\vspace{0mm}\end{eqnarray}}
\newcommand{\barr}{\begin{array}}
\newcommand{\earr}{\end{array}}

\newcommand{\Emat}[0]{{{\bf E}}}

\newcommand{\Imat}{{\bf I}}

\newcommand{\ev}[0]{{\boldsymbol{e}}\xspace}

\newcommand{\hv}[0]{{\boldsymbol{h}}}

\newcommand{\mv}[0]{{\boldsymbol{m}}}

\newcommand{\xv}{\boldsymbol{x}}

\newcommand{\zv}{\boldsymbol{z}}

\newcommand{\thetav}{\boldsymbol{\theta}}

\newcommand{\omegav}[0]{{\boldsymbol{\omega}}}
\newcommand{\R}{\mathbb{R}}
\newcommand{\E}{\mathbb{E}}

\newcommand{\Xcal}{\mathcal{X}}

\newcommand{\Lcal}{\mathcal{L}}
\newcommand{\Ncal}{\mathcal{N}}

\newcommand{\Fcal}{\mathcal{F}}

\newcommand{\Zcal}{\mathcal{Z}}

\ifx\assumption\undefined
\newtheorem{assumption}{Assumption}
\fi

\ifx\definition\undefined
\newtheorem{definition}{Definition}
\fi

\ifx\remark\undefined
\newtheorem{remark}{Remark}
\fi

\usepackage{color, colortbl}
\definecolor{Gray}{gray}{0.93}

\newcommand{\bigblackcircle}{
\begin{tikzpicture}
\filldraw[fill=black,draw=yellow] circle (3pt);
\end{tikzpicture}
}

\usepackage[accepted]{icml2020}

\icmltitlerunning{Feature Quantization Improves GAN Training}

\begin{document}

\twocolumn[
\icmltitle{Feature Quantization Improves GAN Training}



\icmlsetsymbol{equal}{*}

\begin{icmlauthorlist}
\icmlauthor{Yang Zhao}{equal,ub}
\icmlauthor{Chunyuan Li}{equal,msr}
\icmlauthor{Ping Yu}{ub}
\icmlauthor{Jianfeng Gao}{msr}
\icmlauthor{Changyou Chen}{ub}
\end{icmlauthorlist}

\icmlaffiliation{ub}{Department of Computer Science and Engineering, University at Buffalo, SUNY~}
\icmlaffiliation{msr}{Microsoft Research, Redmond}
\icmlcorrespondingauthor{Chunyuan Li}{chunyl@microsoft.com}
\icmlcorrespondingauthor{Changyou Chen}{changyou@buffalo.edu}
\icmlkeywords{GAN}

\vskip 0.3in
]



\printAffiliationsAndNotice{\icmlEqualContribution\\} 

\begin{abstract}
The instability in GAN training has been a long-standing problem despite remarkable research efforts. We identify that instability issues stem from difficulties of performing feature matching with mini-batch statistics, due to a fragile balance between the fixed target distribution and the progressively generated distribution. In this work, we propose Feature Quantization (FQ) for the discriminator, to embed both true and fake data samples into a shared discrete space. The quantized values of FQ are constructed as an evolving dictionary, which is consistent with feature statistics of the recent distribution history. Hence, FQ implicitly enables robust feature matching in a compact space.  
Our method can be easily plugged into existing GAN models, with little computational overhead in training. Extensive experimental results show that the proposed FQ-GAN can improve the FID scores of baseline methods by a large margin on a variety of tasks, including three representative GAN models on 9 benchmarks, achieving new state-of-the-art performance.
\end{abstract}

\section{Introduction}
Generative Adversarial Networks (GANs)~\citep{goodfellow2014generative} are a powerful class of generative models, successfully applied to a variety of tasks such as image generation~\cite{karras2019style}, image-to-image translation~\cite{liu2017unsupervised,zhu2017unpaired,isola2017image}, text-to-image generation~\cite{zhang2017stackgan},  super-resolution~\cite{sonderby2016amortised}, domain adaptation~\cite{tzeng2017adversarial} and sampling from unnormalized distributions~\cite{li2019adversarial}.

Training GANs is a notoriously challenging task, as it involves optimizing a non-convex problem for its Nash equilibrium in a high-dimensional parameter space. In practice, GANs are typically trained via alternatively updating generator and discriminator, using stochastic gradient descent (SGD) based on mini-batches of true/fake data samples. This procedure is often unstable and lacks theoretical guarantees~\cite{salimans2016improved}. Consequently, training may exhibit instability, divergence or mode collapse~\cite{mescheder2018training}. As a result, many techniques to stabilize GAN training have been proposed~\cite{salimans2016improved,miyato2018spectral,karras2019analyzing}.

One possible explanation for the instability is that the learning environment for GANs is non-stationary, and previous models rely heavily on the current mini-batch statistics to match the features across different image regions. Since the mini-batch only provides an estimate, the true underlying distribution can only be learned after passing through a large number of mini-batches. 
This could prevent adversarial learning on large-scale datasets for a variety of reasons: 
$(\RN{1})$ A small mini-batch may not be able to represent true distribution for large datasets, optimization algorithms may have trouble discovering parameter values that carefully search for continuous features to match fake samples with real samples, and these parameterizations may be brittle and prone to failure when applied to previously unseen images. 
$(\RN{2})$ Increasing the size of the mini-batch can increase the estimation quality, but doing this also loses the computational efficiency obtained by using SGD.
$(\RN{3})$ In particular, the distribution of fake samples shifts as the generator changes during training, making the classification task for discriminator evolve over time~\cite{chen2019self,liang2018generative,zhao2020leveraging,cong2020lmemory}. In such a non-stationary online environment, discriminator can forget previous tasks if it relies on the statistics from the current single mini-batch, rendering training unstable. 



\begin{figure*}[t!]
	\vspace{-0mm}\centering
	\begin{tabular}{c}
		\hspace{-0mm}
		\includegraphics[height=3.7cm]{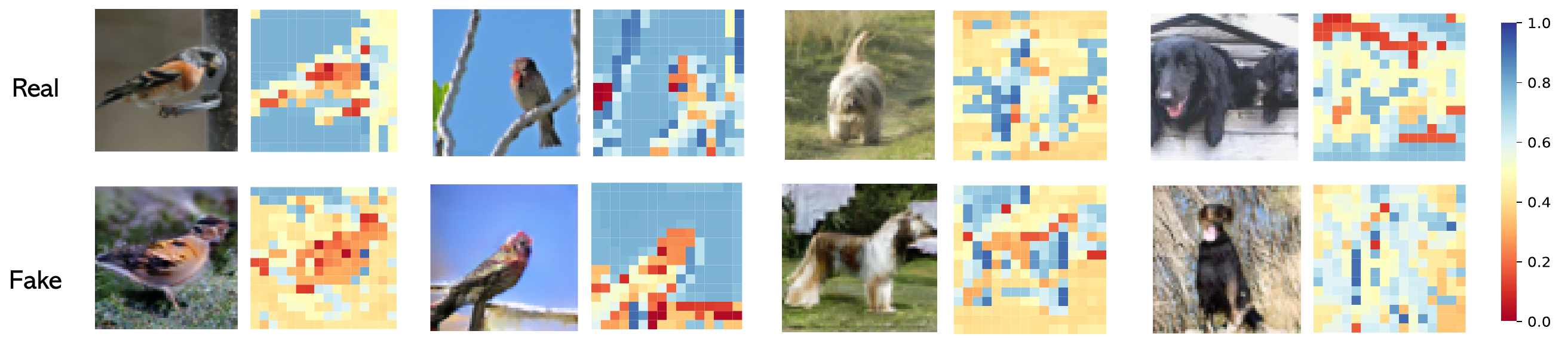} 
	\end{tabular}
	\vspace{-2mm}
	\caption{
	The proposed FQ-GAN generates images by leveraging quantized features from a dictionary, rather than producing arbitrary features in a continuous space when judged by the discriminator. The odd columns show images of the same class (real on the top row, fake at the bottom row), whose corresponding quantized feature maps are shown in the right even column, respectively. The dictionary items are visualized in 1D as the color-bar using t-SNE~\cite{maaten2008visualizing}. 
	Image regions with similar semantics utilize the same/similar dictionary items. For example, bird neck is in dark red, sky or clear background is in shallow blue, grass is in orange. 
	 }
	\vspace{-2mm}
	\label{fig:fq_img_dict}
\end{figure*}

In this work, we show that GANs benefit from feature quantization (FQ) in the discriminator. A dictionary is first constructed via moving-averaged summary of features in recent training history for both true and fake data samples. This enables building a large and consistent dictionary on-the-fly that facilitates the online fashion of GAN training. Each dictionary item represents a unique feature prototype of similar image regions. By quantizing continuous features in traditional GANs into these dictionary items, the proposed FQ-GAN forces true and fake images to construct their feature representations from the limited values, when judged by discriminator. This alleviates the poor estimate issue of mini-batches in traditional GANs.


To better understand what has been learned during the generation process, we visualize the quantized feature maps of the discriminator in FQ-GAN for different images. Some sample images are shown in Figure~\ref{fig:fq_img_dict}. Image regions with similar semantics utilize the same or similar dictionary items.

The contributions of this paper are summarized as follows:
$(\RN{1})$
We propose FQ, a simple yet effective technique that can be added universally to yield better GANs.
$(\RN{2})$
The effectiveness of FQ is validated with three GAN models on 10 datasets. Compared with traditional GANs, we show empirically that the proposed FQ-GAN helps training converge faster, and often yields performance improvement by a large margin, measured by generated sample quality. 
The code is released on Github\footnote{\url{https://github.com/YangNaruto/FQ-GAN}}.

\section{Background}

\subsection{Preliminaries on vanilla GANs}
Consider two general marginal distributions $q(\xv)$ and $p(\zv)$ over $\xv \in \Xcal $ and $\zv \in \Zcal$.
To generate samples from these random variables, adversarial methods \cite{goodfellow2014generative} provide a sampling mechanism that only requires gradient backpropagation, without the need to specify the conditional densities.
Specifically, instead of sampling directly from the desired conditional distribution, the random variable is generated as a deterministic transformation of an independent noise, \eg a Gaussian distribution. 
The sampling procedure for conditionals $\tilde{\xv} \sim p_{\thetav}(\xv | \zv)$ is carried out through the following generating process:
\begin{align}\label{eq:conditional_sample}
\tilde{\xv} = g_{\thetav}(\zv), \ \zv \sim p(\zv),
\end{align}
where $g_{\thetav}(\cdot)$ is the generators, specified as neural networks with parameters $\thetav$, 
and $p(\zv)$ is specified as a simple parametric distribution, \eg isotropic Gaussian  $p(\zv) = \Ncal(\zv; 0,\Imat)$. 
Note that \eqref{eq:conditional_sample} implies that $p_{\thetav}(\xv | \zv)$ is parameterized by $\thetav$, hence the subscripts.

The goal of GAN~\cite{goodfellow2014generative} is to match the marginal
$p_{\thetav}(\xv) = \int p_{\thetav}(\xv | \zv) p(\zv){\rm d}\zv$ to $q(\xv)$.
Note that $q(\xv)$ denotes the true distribution of the data, from which we have samples.
In order to do the matching, GAN trains a $\omegav$-parameterized adversarial discriminator network, $f_{\omegav}(\xv)$, to distinguish between samples from $p_{\thetav}(\xv)$ and $q(\xv)$.
Formally, the minimax objective of GAN is given by the following expression:
\begin{align}\label{eq:gan}
 \min_{\thetav} \max_{\omegav} \ \
& \Lcal_{\rm GAN} =  
 \E_{\xv \sim q(\xv)} [ \log \sigma (f_{\omegav}(\xv) )] + \nonumber \\
& \E_{\tilde{\xv} \sim p_{\thetav}(\xv| \zv ),\zv \sim p(\zv)} [ \log (1-\sigma( f_{\omegav}(\tilde{\xv}) ) ) ],
\end{align}
where $\sigma(\cdot)$ is the sigmoid function.

\begin{figure*}[t!]
	\vspace{-0mm}\centering
	\begin{tabular}{c c}
		\hspace{-0mm}
		\includegraphics[height=3.0cm]{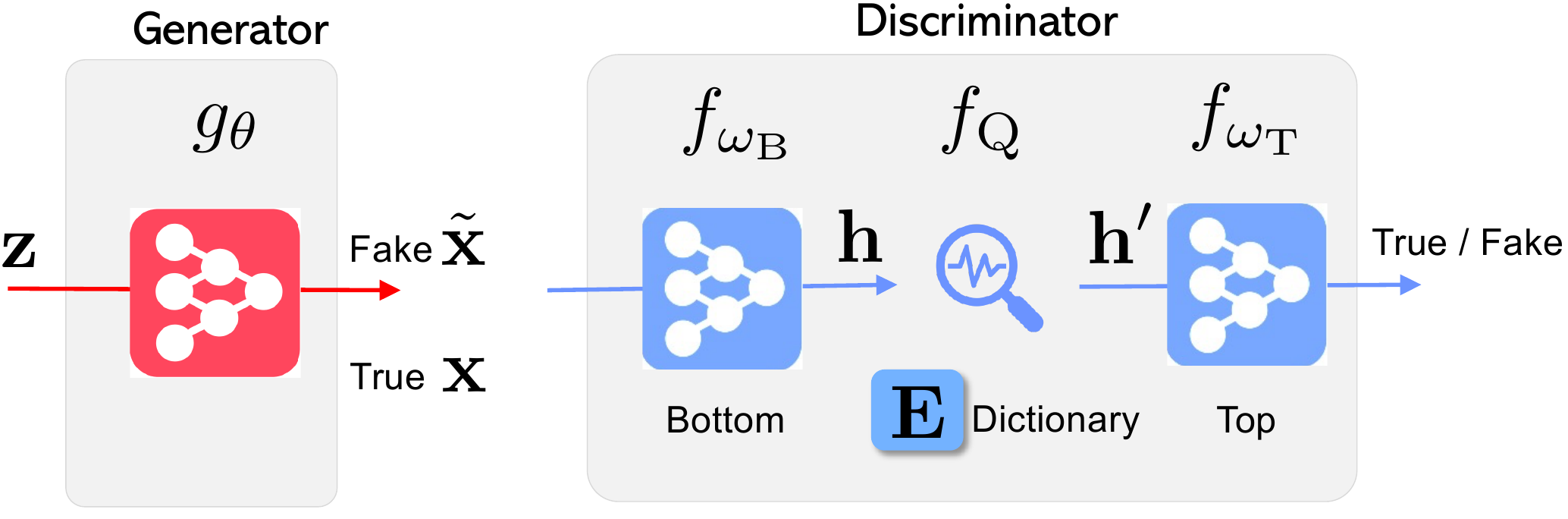}  & 
		\includegraphics[height=3.5cm]{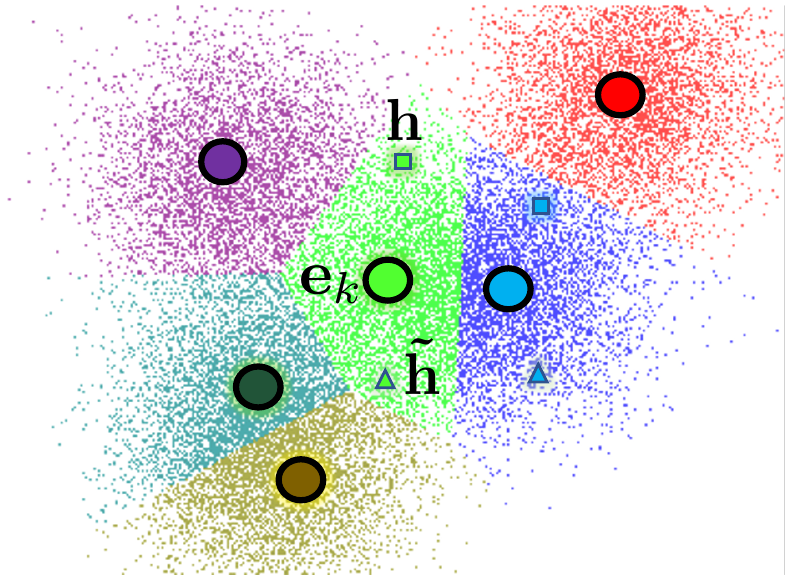} \\
		(a) FQ-GAN architecture \vspace{2mm} & 
		(b) Dictionary look-up \hspace{-0mm} \\ 
	\end{tabular}
	\vspace{-2mm}
	\caption{Illustration of FQ-GAN: (a) The neural network architecture. A feature quantization (\ie dictionary look-up) step $f_{\rm Q} $ is injected into the discriminator of the standard GANs. (b) A visualization example of dictionary $\Emat$ and the look-up procedure. Each circle ``$\bigblackcircle$'' indicates a quantization centroid. The true sample features $\hv$ (``$\blacksquare$'') and fake sample features $\tilde{\hv}$ (``$\blacktriangle$'') are quantized into their nearest centroids $\ev_k$ (represented in the same color in this example), and thus performing implicit feature matching.
	 }
	\vspace{-2mm}
	\label{fig:fq_gan_illustration}
\end{figure*}

\subsection{Pitfall of Continuous Features}

Several works have shown that using feature matching as a training objective of GANs
can improve model performance. The basic idea is to embed true/fake distributions in a finite-dimensional continuous feature space, and to match them based on their feature statistics using some divergence metrics. One general form of feature matching  is based on Integral probability metric (IPM)~\cite{muller1997integral}, indexed by the function space $\Fcal$, defined as follows:
\begin{align}\label{eq:ipm}
d_{\Fcal} (p, q) = \sup_{f \in \Fcal} 
| \E_{\tilde{\xv} \sim p(\tilde{\xv}) } f(\tilde{\xv}) - E_{\xv \sim q(\xv)} f(\xv) |
\end{align}
The particular function class $\Fcal$ determines the probability metric. For example, \citet{mroueh2017mcgan} proposed MC-GAN, which utilizes both mean and covariance feature statistics. They further showed that several previous works on GAN can be written within the mean feature matching framework, including Wasserstein GAN~\cite{arjovsky2017wasserstein}, MMD-GAN~\cite{li2017mmd}, and Improved GAN~\citep{salimans2016improved}. 

Though theoretically attractive, these continuous feature matching methods fall short of recent state-of-the-art performance on large datasets~\cite{brock2018large,karras2019style}. We argue there are two issues: 
$(\RN{1})$ Principled methods often require to constrain the discriminator capacity (\eg weight clipping or gradient penalty) to ensure the boundedness: new architectural adjustments such as a higher number of feature maps are needed to compensate for constraints~\cite{mroueh2017mcgan}. The question remains what architectural choices can balance the trade-off in practice.
$(\RN{2})$ More importantly, the direct feature matching scheme in~\eqref{eq:ipm} is estimated via mini-batch statistics, which can be prohibitively inaccurate on large or complex datasets. An effective alternative to match features at large-scale is required, even if it is indirect. 



\section{Feature Quantization GANs}

\subsection{From Continuous to Quantized Representations}
Without loss of generality, the discriminator $f_{\omegav}(\xv)$ can be rewritten with a function decomposition: 
\begin{align}\label{eq:discriminator_2func}
 f_{\omegav}(\xv) = f_{\omegav_T} \circ f_{\omegav_B}(\xv),
\end{align}
where $f_{\omegav_B}(\xv)$ is the bottom network, whose output feature $\hv \in \R^{D}$ is in a $D$-dimensional continuous space, and used as the input of the top network $f_{\omegav_T}(\hv)$.
Instead of working in a continuous feature space, we propose to quantize features into a discrete space, enabling implicit feature matching. 


Specifically, we consider the discrete feature space as a dictionary $\Emat = \{ \ev_k \in \R^D  \mid k \in 1, 2, \cdots, K \}$, where $K$ is the size of the discrete space (\ie a $K$-way categorical space), and $D$ is the dimensionality of each dictionary item $\ev_k$.

The discrete feature $\hv^{\prime}$ is then calculated by a nearest neighbour look-up using the shared dictionary:
\begin{align}\label{eq:quantization_nn}
 \hv^{\prime}  = f_{\rm Q}  (\hv)= \ev_k, ~\text{where}~~
  k = \arg\!\min_j \| \hv  - \ev_j \|_2~,
\end{align}
where $f_{\rm Q}$ is a parameter-free look-up function, and $\hv^{\prime}$ is further sent to the top network. Hence, in contrast to the traditional discriminator in~\eqref{eq:discriminator_2func}, our feature quantization discriminator is:   
\begin{align}\label{eq:discriminator_3func}
 f_{\omegav}(\xv) = f_{\omegav_T} \circ  f_{\rm Q} \circ f_{\omegav_B}(\xv),
\end{align}
The overall scheme of FQ-GAN is illustrated in Figure~\ref{fig:fq_gan_illustration}. 
FQ-GAN in \eqref{eq:discriminator_3func} reduces to the standard GAN model in \eqref{eq:discriminator_2func} if $f_{\rm Q}$ is removed.

%


\subsection{Dictionary Learning}
One remaining question is how to construct the dictionary $\Emat$. Following~\citep{van2017neural}, we consider a feature quantization loss consisting of two terms specified in \eqref{eq:quantization_loss}: 
$(\RN{1})$ The {\it dictionary loss}, which only applies to the dictionary items, brings the selected item $\ev$ close to the output of the bottom network. 
$(\RN{2})$ The {\it commitment loss} encourages the output of the bottom network to stay close to the chosen dictionary item to prevent it from fluctuating too frequently from one code item to another. The operator $\mathtt{sg}$ refers to a {\it stop-gradient} operation that blocks gradients from flowing into its argument, and $\beta$ is a weighting hyper-parameter ($\beta = 0.25$ in all our experiments):
\begin{align}\label{eq:quantization_loss}
\hspace{-2mm}
\Lcal_{\rm Q} 
\!=\!  \underbrace{\| \mathtt{sg}(\hv) - \ev_k  \|_2^2}_{\text{dictionary loss}} 
+ \beta \underbrace{ \| \mathtt{sg}(\ev_k) - \hv  \|_2^2}_{\text{commitment loss}}~, 
\end{align}
where $\ev_k$ is the nearest dictionary item to $\hv$ defined in~\eqref{eq:quantization_nn}. 

\paragraph{A dynamic \& consistent dictionary} The evolution of the generator during GAN training poses a continual learning problem for the discriminator~\cite{liang2018generative}. In another word, the classification tasks for the discriminator change over time, and recent samples from the generator are more related to current discriminator learning.
This inspires us to maintain the dictionary as a queue of features, allowing reusing the encoded features from the preceding mini-batches. The current mini-batch is enqueued to the dictionary, and the oldest mini-batches in the queue are gradually removed. The dictionary always represents a set of prototypes for the recent features, while the extra computation of maintaining this dictionary is manageable. Moreover, removing the features from the oldest mini-batch can be beneficial, because its encoded features are from an early stage of GAN training, and thus the least realistic and consistent with the newest ones. 

Alternatively, one may wonder learning a dictionary using all training data beforehand, and keep the dictionary fixed during GAN training. We note this scheme is not practical in that 
$(\RN{1})$ Modern datasets such as ImageNet are usually very large, learning a dictionary offline is prohibitively computationally expensive.  
$(\RN{2})$ More importantly, such a dictionary is not representative for fake images at the early of training, rendering it difficult to effectively learn quantized features for fake images.


\begin{algorithm}[!t]
   \caption{Feature Quantization GAN}
   \label{alg:example}
\begin{algorithmic}
    \REQUIRE Randomly initializing the parameters of generator $g_{\thetav}$, discriminator $f_{\omegav}$, and dictionary $\Emat$
   \FOR{a number of training iterations}
   
        \STATE {\quad {\small \color{blue} $\#  \mathtt{~Produce~a~mini batch~ of~true~and~fake~samples~}$}}    
        \STATE {\quad Sample $\zv \sim p(\zv)$ and true samples  $\xv \sim q(\xv)$};
        \STATE {\quad Forward $\zv$ to generate fake samples  $\tilde{\xv} = g_{\thetav}(\zv)$};
        
        \STATE{\quad {\small \color{blue} $\# \mathtt{~Feature~~quantization~\&~Dictionary~ learning}$}}   
        \STATE{\quad Forward samples $\{\xv, \tilde{\xv}\}$ using~\eqref{eq:discriminator_3func}, and produce $\hv$};
        \STATE {\quad Feature quantization using~\eqref{eq:quantization_nn}};
        \STATE {\quad Momentum update of dictionary $\Emat$ using~\eqref{eq:momentum_update}};
        
        \STATE{\quad {\small \color{blue} $\# \mathtt{~Update~discriminator}$}} 
        \STATE {\quad Compute gradient $\frac{\partial \Lcal_{\rm FQGAN} }{\partial \omegav }$ of~\eqref{eq:fqgan}};
        \STATE {\quad Update $\omegav$ via gradient ascent};
        
        \STATE{\quad {\small \color{blue} $\# \mathtt{~Update~generator}$}} 
        \STATE {\quad Compute gradient $\frac{\partial \Lcal_{\rm FQGAN} }{\partial \thetav }$ of~\eqref{eq:fqgan}};
        \STATE {\quad Update $\thetav$ via gradient descent};
   \ENDFOR
\end{algorithmic}
\end{algorithm}

\paragraph{Momentum update of dictionary.}
Specifically, we use the exponential moving average updates to implement the evolving dictionary, as a replacement for the dictionary loss term in~\eqref{eq:quantization_loss}. For a mini-batch of size $n$, $n_k$ is the number of features that will be quantized to dictionary item $\ev_k$. The momentum update for $\ev_k$ is:
\begin{align}\label{eq:momentum_update}
\ev_k \leftarrow \mv_k /N_k, ~\text{where}~~ 
\mv_k \leftarrow~ & \lambda  \mv_k + (1-\lambda) \sum_{i=1}^{n_k} \hv_{i,k}, \nonumber \\
N_k \leftarrow~ & \lambda  N_k + (1-\lambda) n_k ,
\end{align}
where $\lambda \in (0, 1)$ is a momentum coefficient. Only the parameters in the bottom network $f_{\omegav_B}$ are updated by back-propagation. The momentum updates above make $\ev_k$ evolve more smoothly. Small $\lambda$ considers less history. For example, $\lambda=0$ only utilizes the current mini-batch statistics and ignores the entire history, thus \eqref{eq:momentum_update} reduces to~\eqref{eq:quantization_loss}. We used the default $\lambda = 0.90$ in all our experiments. 


\vspace{-2mm}
\subsection{FQ-GAN Training}

The overall training objective of the proposed FQ-GAN is:
\begin{align}\label{eq:fqgan}
 \min_{\thetav,~\Emat} \max_{\omegav} \ \
& \Lcal_{\rm FQ-GAN} = \Lcal_{\rm GAN} +  \alpha \Lcal_{\rm Q},
\end{align}
where $\alpha$ is the weight to incorporate the proposed FQ into GANs. The training procedure is detailed in Algorithm 1. In practice, to avoid degeneration, $\alpha$ can be annealed from 0 to 1, and $\hv$ (instead of $\hv^{\prime}$) can be used to feed to the next layer at the beginning of training. In this case, one may consider FQ regularizes the learned features using clustering. The generator parameter $\thetav$ and discriminator parameter $\omegav$ are updated via the regularized GAN objective in~\eqref{eq:fqgan},
while the dictionary items $\Emat$ are updated via~\eqref{eq:momentum_update}.   
FQ-GAN enjoys several favorable properties, explained as follows. 

\paragraph{Scalability.}
The introduction of a dynamic dictionary decouples the dictionary size from the mini-batch size. The dictionary size can be much larger than a typical mini-batch size, and can be flexibly and independently set as a hyper-parameter. The items in the dictionary are progressively replaced. Compared with traditional feature matching methods that only consider the current mini-batch statistics, the proposed FQ-GAN maintains much more representative feature statistics in the dictionary, allowing robust feature matching for large datasets.

\paragraph{Implicit feature matching.} FQ-GAN shares similar spirits of many other regularization techniques for the discriminator in that they reduce the representational power of the discriminator. However, instead of imposing boundness on weights or gradients, FQ-GAN restricts continuous features into a prescribed set of values, \ie feature centroids. Since both true and fake samples can only choose their representations from the limited dictionary items, FQ-GAN indirectly performs feature matching. This can be illustrated using the visualization example in Figure~\ref{fig:fq_gan_illustration} (b), where true features $\hv$ and fake features $\tilde{\hv}$ are quantized into the same centroids. Further, the discrete nature improves the possibilities of feature matching, compared to a continuous space.


\begin{figure}[t!]
	\vspace{-0mm}\centering
	\begin{tabular}{c}
		\hspace{-3mm}
		\includegraphics[height=2.7cm]{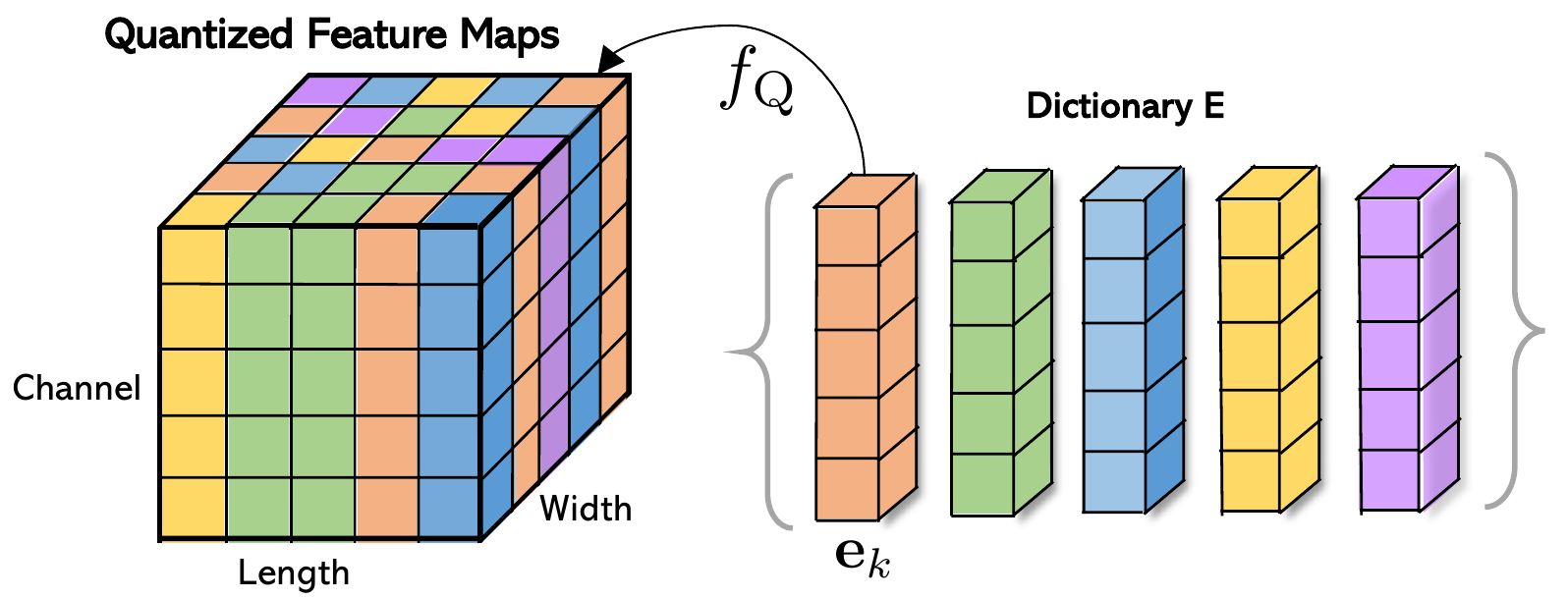} 
	\end{tabular}
	\vspace{-2mm}
	\caption{Illustration of FQ construction in CNNs. In this example, the dictionary has 5 items, and feature map is $\hv \in {\R^{5\times 5 \times 5}}$. The feature vector at each position is quantized into a dictionary item, \eg the back-right feature is quantized into a red item.
	 }
	\vspace{-2mm}
	\label{fig:fq_cnn_schemes}
\end{figure}

\vspace{-2mm}
\subsection{FQ-GAN for image generation}
FQ is a general method for discriminator design in GANs. We consider image generation tasks in this paper, where the discriminator is often parameterized by convolutional neural networks (CNNs). Each image is represented as a feature map $\hv \in {\R^{C\times L \times W}}$ in CNNs, where $C$, $L$, $W$ is the number of channels as well as the length and width, respectively. We construct a position-wise dictionary, with each item $\ev \in {\R^{C}}$. At a given position on the feature map, the feature vector characterizes the local image region. It is quantized into its nearest dictionary item for calibration, leading to a new quantized feature map $\hv^{\prime}$ containing calibrated local feature prototypes.
We provide the visual illustration on constructing FQ for CNN-based discriminator in Figure~\ref{fig:fq_cnn_schemes}. Note that the FQ module can be used in multiple different layers of discriminator.




%


\vspace{-1mm}
\section{Related Work}
\vspace{-0mm}

\subsection{Improving GANs}
Training vanilla GANs is difficult: it requires carefully finely-tuned hyper-parameters and network architectures to make it work. Much recent research has accordingly focused on improving its stability, drawing on a growing body of empirical and theoretical insights~\cite{nowozin2016f,li2017alice,zhu2017unpaired,fedus2017many}. Among them, the three following aspects are related to FQ-GAN. 
\paragraph{Regularized GANs.} Various Regularization methods have been proposed, including changing the objective functions to encourage convergence~\cite{arjovsky2017wasserstein,mao2017least,mescheder2018training,kodali2017convergence,zhang2019consistency}, and constraining discriminator through gradient penalties ~\cite{gulrajani2017improved} or normalization~\cite{miyato2018spectral}. They counteract the use of unbounded loss functions and ensure that discriminator provides gradients everywhere to generator.
FQ is also realted to variational discriminator bottleneck~\cite{peng2018variational} in the sense that both restrict the feature representation capacity.
BigGANs~\cite{brock2018large} use orthogonal regularization, and achieve  state of the art image synthesis performance on ImageNet~\cite{imagenet_cvpr09}.

\paragraph{Network architectures.} Recent advances consider architecture designs, such as SA-GAN~\cite{zhang2019self}, which adds the self-attention block to capture global structures.  Progressive-GAN~\cite{karras2018progressive} trains high-resolution GANs in the single-class setting by training a single model across a sequence of increasing resolutions. As a new variant, Style-GAN~\cite{karras2019style} proposed a generator architecture to separate high-level attributes and stochastic variation, achieving highly varied and high-quality human faces.

\paragraph{Memory-based GANs.} 
\citet{kim2018memorization} increased the model complexity via proposing a shared and sophisticated memory module for both generator and discriminator. The generation process is conditioned on samples from the memory. ~\citet{zhu2019dm} proposed a dynamic memory component for image refinement in the text-to-image task.


Compared with the above three aspects, 
FQ-GAN slightly modifies the discriminator architecture by injecting a dictionary-based look-up layer, and thus regularizes the model capacity to encourage easier feature matching. The dictionary in FQ-GAN can be viewed as a much simpler memory module to store feature statistics.
Importantly, our FQ-GAN is easier to use and orthogonal to existing GANs, and can be simply employed as a plug-in module to further improve their performance.

\subsection{Vector Quantization}

Vector quantization (VQ)~\citep{gray1984vector} has been used in various settings, including clustering~\cite{equitz1989new}, metric learning~\cite{schneider2009distance}, etc.
The most related work to ours is~\citep{van2017neural}, where discrete latent representations are proposed for variational auto-encoders to circumvent the issues of ``posterior collapse'', and show that pairing such quantized representations with an autoregressive prior can generate high-quality images, videos, and speech. 
Our motivation and scenarios are different from previous VQ works. To the best of our knowledge, this paper presents the first feature quantization work for GANs. 

\vspace{-2mm}
\section{Experiments}

%


We apply the proposed FQ-GAN method to three state-of-the-art GAN models for a variety of tasks. 
$(\RN{1})$ 
BigGAN~\citep{brock2018large} for image synthesis, especially for ImageNet, representing a generation task for large-scale datasets.
$(\RN{2})$ 
StyleGAN~\citep{karras2019style,karras2019analyzing} for face synthesis, representing a generation task for high-resolution images.
$(\RN{3})$ 
U-GAT-IT~\cite{kim2019u} for an unsupervised image-to-image translation task.
\begin{figure*}[t!]
	\vspace{-0mm}\centering
	\begin{tabular}{c c c c}
		\hspace{-3mm}
		\includegraphics[height=2.7cm]{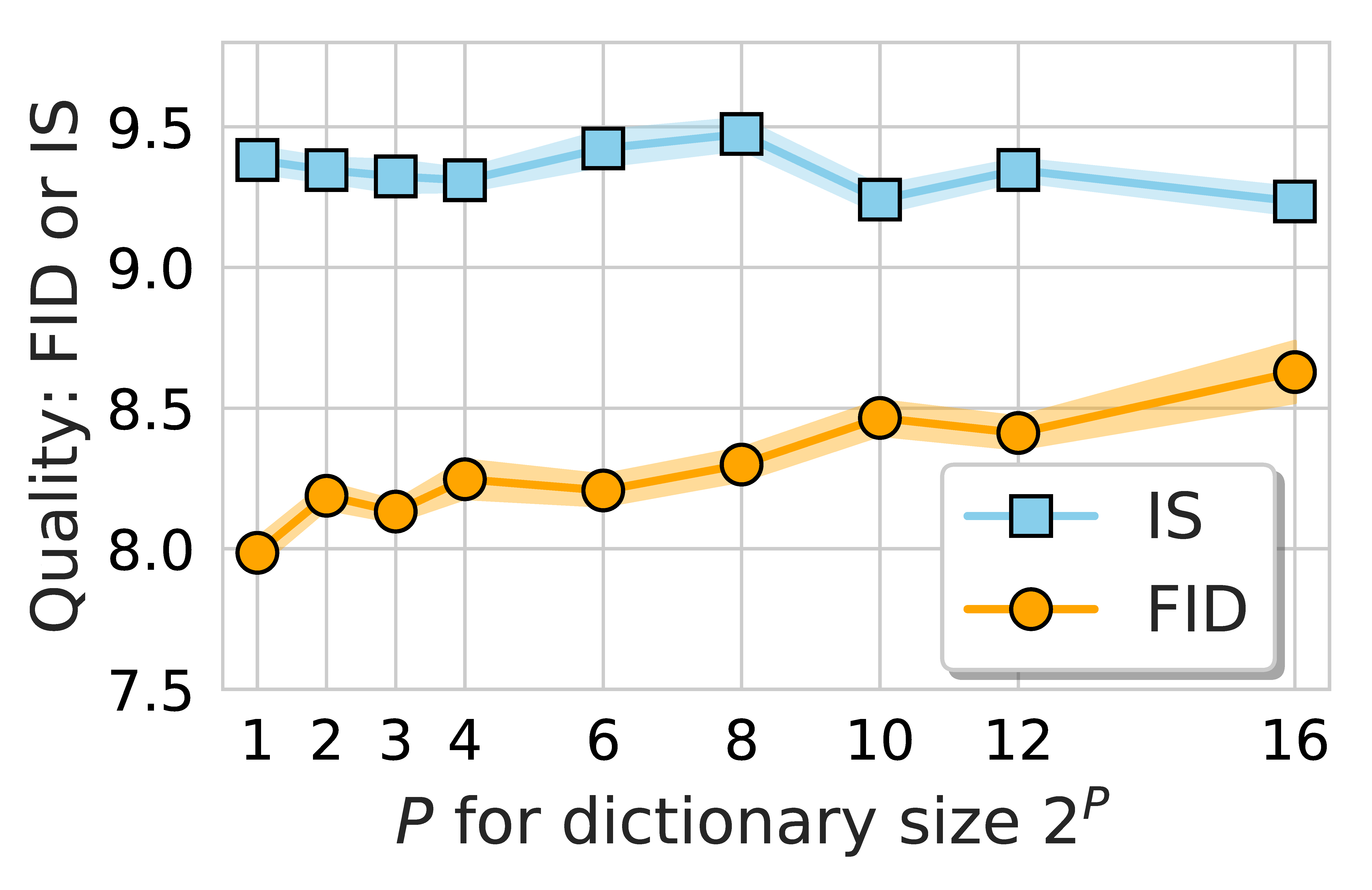}  & 
		\hspace{-4mm}
		\includegraphics[height=2.7cm]{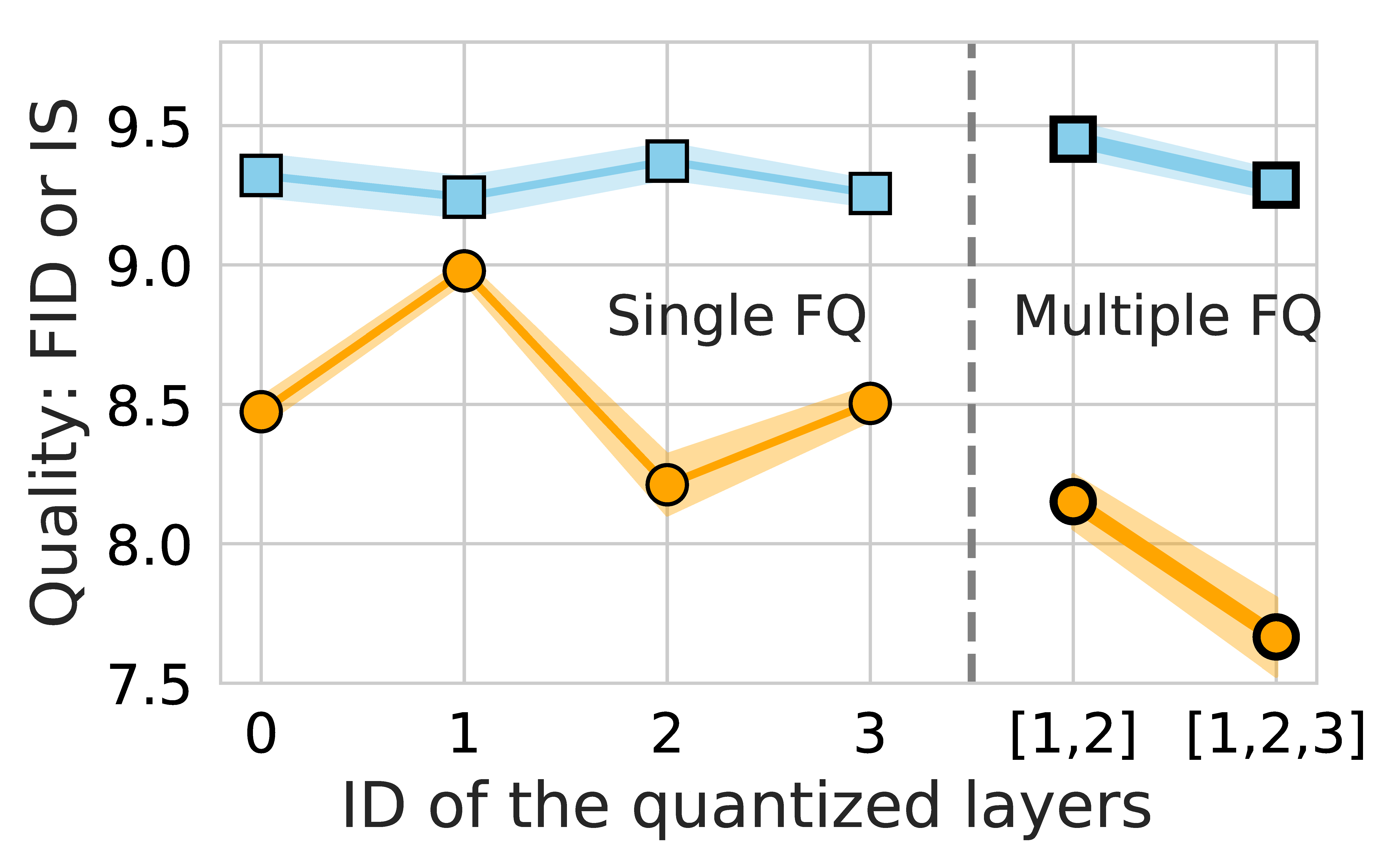}  &
		\hspace{-4mm}
		\includegraphics[height=2.7cm]{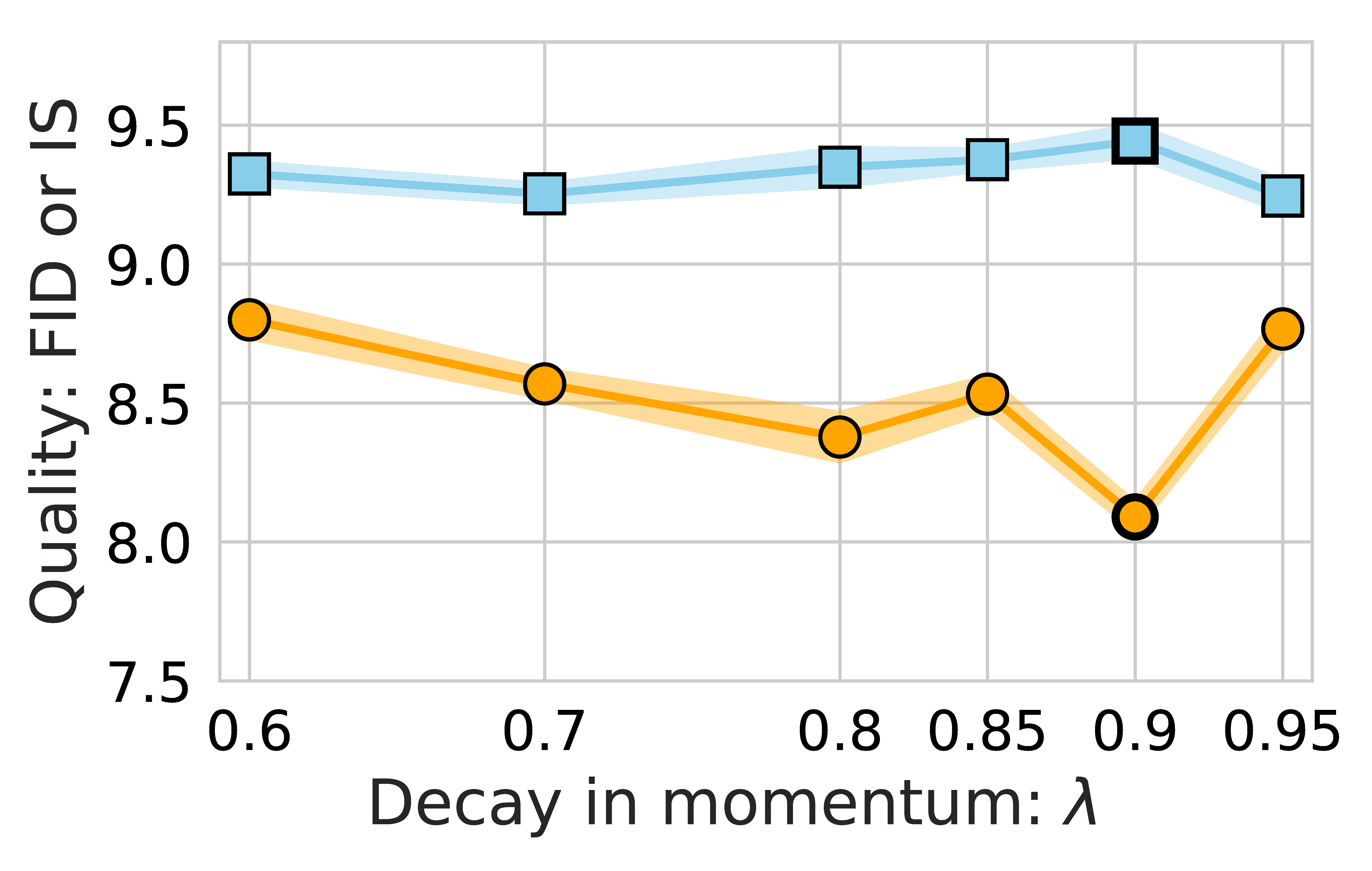}  &
		\hspace{-4mm}
		\includegraphics[height=2.7cm]{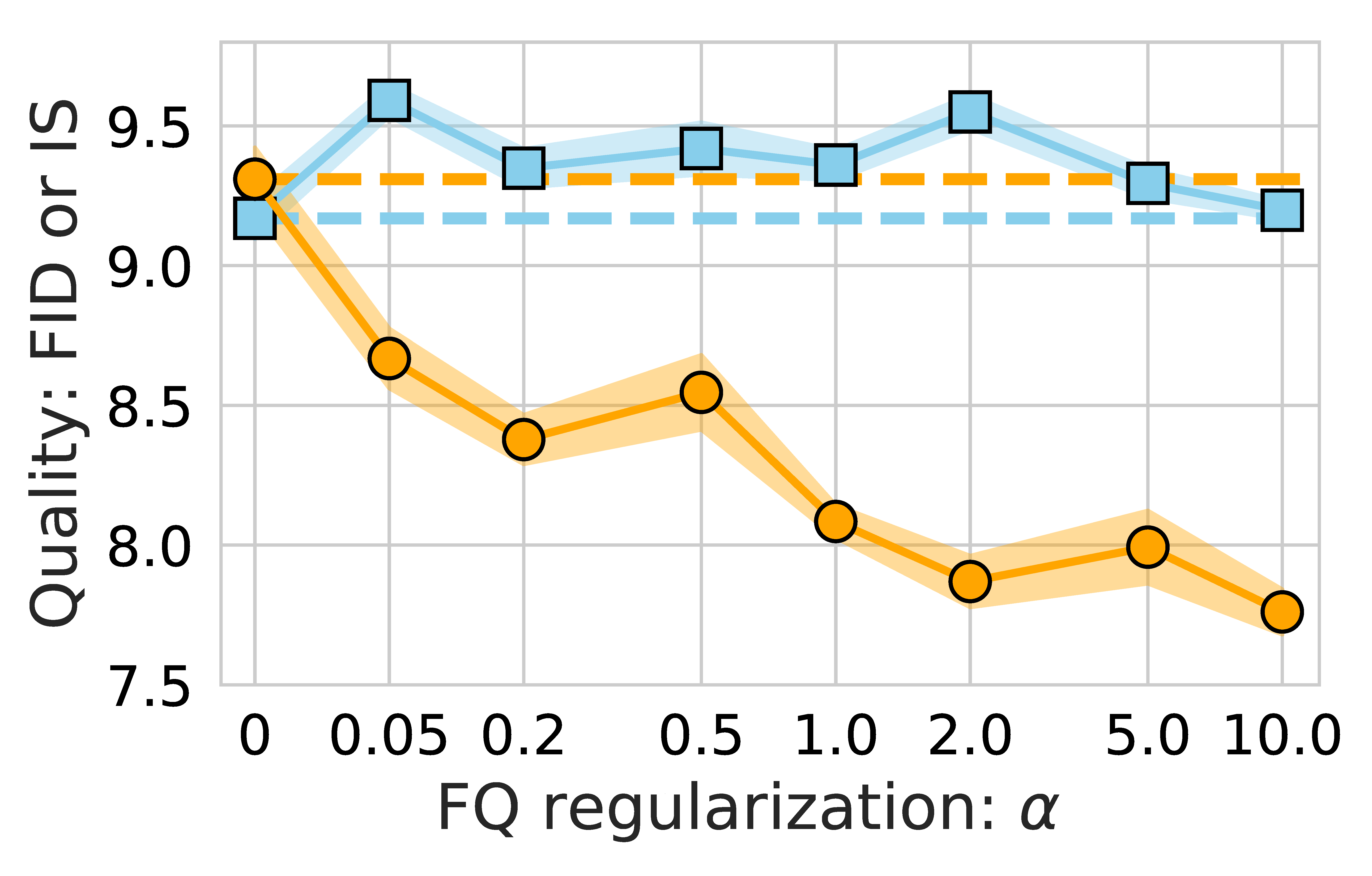} \\
		(a) Dictionary size \vspace{0mm} & 
		(b) FQ layer position \hspace{-0mm} & 
		(c) $\lambda$ \hspace{0mm} & 
		(d) $\alpha$ \hspace{-0mm} \\ 
	\end{tabular}
	\vspace{-3mm}
	\caption{Ablation studies on the impact of hyper-parameters. The image generation quality is measured with FID $\downarrow$ and IS $\uparrow$. (a) Dictionary size $K=2^P$. (b) The positions to apply FQ to discriminator, layer ID is shown on the horizontal axis. (c) The decay hyper-parameter $\lambda$ in dictionary update. (d) The weight $\alpha$ to incorporate FQ, the dashed horizon lines are standard GAN baseline $\alpha = 0$.
	 }
	\vspace{-4mm}
	\label{fig:ablation}
\end{figure*}

\paragraph{Evaluation metrics.}
 We consider three commonly used evaluation metrics for GANs.
$(\RN{1})$ 
    {\it Inception Score (IS)}~\citep{salimans2016improved}  measures how realistic the output of the generator is and the intra-class variety, based on how well the image classification model Inception v3 \citep{szegedy2016rethinking} classifies them as one of $1,000$ known objects collected in ImageNet-1000. Higher scores mean that the model can generate more distinct images. However, it is not reliable when generated images concentrate to the class centers.
$(\RN{2})$ 
    {\it Fréchet Inception Distance (FID)}~\citep{heusel2017gans} compares the statistics (mean and variances of Gaussian distributions) between the generated samples and real samples. FID is consistent with increasing disturbances and human judgment. Lower scores indicate that the model can generate higher quality images.
$(\RN{3})$ 
    {\it Kernel Inception Distance (KID)}~\citep{binkowski2018demystifying} improves FID as an unbiased estimator, making it more reliable when there are fewer available test images. We use generated images translated from all test images in the source domain vs. test images in the target domain to compute KID. Lower KID values indicate that images are better translated.

All the baseline methods are implemented via the official codebases from the authors. Model variants that incorporate the proposed feature quantization technique are named with prefix ``\textbf{\texttt{FQ}}''.  Experiment details are provided in Appendix.


\vspace{-2mm}
\subsection{On the impact of hyper-parameters}
We investigate the hyper-parameters of FQ on the CIFAR-100 dataset~\citep{krizhevsky2009learning} . It has 100 classes containing 600 images each, in which there are 500 training images and 100 testing images. Four-layer networks are employed for both the generator and the discriminator. We train the model for 500 epochs, and save a model every 1000 iterations.  We take the last 10 checkpoints to report the mean of their performance, measured by FID and IS.

\paragraph{Dictionary size $K$.} In Figure~\ref{fig:ablation} (a), we show the FQ-GAN performance with various dictionary size $K=2^P$. We see that a smaller $K$ yields better performance. Surprisingly, the dictionary with binary values $K=2~(P=1)$ provides the best results on this dataset. 
Larger $K$ is less favorable for two reasons: (1) it can be more memory expensive; (2) the method becomes similar to the continuous feature variant, and $K \rightarrow \infty$ recovers original GANs.  Hence, we suggest to choose a smaller $K$ when the performances are similar. 


\paragraph{Which player/layer to add FQ?} The proposed FQ module can be plugged into either a generator or a discriminator. We found that the performance does not change much when used in the generator. For example, FID is 9.01 {\scriptsize $\pm .44$} and 8.96 {\scriptsize $\pm .26$} before and after adding FQ, respectively. For the discriminator, we place FQ at different positions of the network, and show the results in Figure~\ref{fig:ablation} (b). Multiple FQ layers can generally outperform a single FQ layer. 


\paragraph{Momentum decay $\lambda$.} Note that $\lambda$ determines how much recent history to incorporate when constructing the dictionary. Larger values consider more history. Our experimental results in Figure~\ref{fig:ablation} (c) show that $\lambda=0.9$ is a sweet point to balance the current and historical statistics. 

\paragraph{FQ weight $\alpha$.} The impact of weighting hyper-parameter $\alpha$ for FQ in \eqref{eq:fqgan} is studied in Figure~\ref{fig:ablation} (d). Adding FQ can immediately improve the baseline by a large margin. Larger $\alpha$ can further decrease FID while keeping IS values almost unchanged. We used $\alpha=1$ for convenience.

\begin{table}[!t]
    \centering
    \begin{tabular}{@{}c|c|c@{}}
    \toprule
   Model &   \!\texttt{FID}* $\downarrow$ / \!\texttt{IS}* $\uparrow$  & \!\texttt{FID} $\downarrow$ / \!\texttt{IS} $\uparrow$      \\ \midrule
          SN-GAN    & 14.26 / 8.22 & $-$    \\
          R-MMD-GAN  &  ~16.21 / 8.29$^\dagger$  &  $-$  \\ \hline
          BigGAN    & 6.04 / 8.43  &    6.30{\scriptsize $\pm$.20} / 8.31{\scriptsize $\pm$.12}  \\
          \rowcolor{Gray} \textbf{\texttt{FQ}}-BigGAN  &  
          \textcolor{blue}{\textbf{5.34}} / \textcolor{blue}{\textbf{8.50}}   & 
          \textcolor{blue}{\textbf{5.59{\scriptsize$\pm$.12}} }/ 
          \textcolor{blue}{\textbf{8.48{\scriptsize$\pm$.03}} } \\
    \bottomrule
    \end{tabular}
     \vspace{-2mm}
    \caption{Comparison on CIFAR-10. $^\dagger$This number is quoted from~\citep{wang2019improving}}
    \vspace{-2mm}
    \label{tab:score_cifar10}
\end{table}
\begin{table}[!t]
    \centering
    \begin{tabular}{c|c|c}
    \toprule
   Model &  \texttt{FID}* $\downarrow$ / \texttt{IS}* $\uparrow$  & \texttt{FID} $\downarrow$ / \texttt{IS} $\uparrow$     \\ \midrule
          SN-GAN    & 16.77 / 7.01 &  $-$  \\ \hline
          TAC-GAN      & ~7.22 / 9.34$^\dagger$ & $-$  \\
          \rowcolor{Gray} 
          \textbf{\texttt{FQ}}-TAC-GAN  &  \textcolor{blue}{\textbf{7.15}} / \textcolor{blue}{\textbf{9.74}} & 
          \textcolor{blue}{\textbf{7.21{\scriptsize $\pm$.10}}} / \textcolor{blue}{\textbf{9.69{\scriptsize $\pm$.04}}} \\ \hline
          BigGAN      & 8.64 / 9.46 & 9.01{\scriptsize $\pm$.44} / 9.36{\scriptsize $\pm$.10}  \\
          \rowcolor{Gray} 
          \textbf{\texttt{FQ}}-BigGAN  &  \textcolor{blue}{\textbf{7.36}} / \textcolor{blue}{\textbf{9.62}} & 
          \textcolor{blue}{\textbf{7.42{\scriptsize $\pm$.07}}} / \textcolor{blue}{\textbf{9.59{\scriptsize $\pm$.04}}} \\
    \bottomrule
    \end{tabular}
     \vspace{-2mm}
    \caption{Comparison on CIFAR-100. $^\dagger$This number is quoted from~\citep{gong2019twin}.}
    \label{tab:score_cifar100}
     \vspace{-2mm}
\end{table}

\begin{table}[t!]
    \centering
  \vspace{-0mm}
    \begin{minipage}{0.5\textwidth}
    \begin{tabular}{@{}c@{}c@{}|c|@{}c}
    \toprule 
    \multicolumn{2}{c|}{\multirow{2}{*}{Models}}
    & $64\times64$ & $128\times128$ \\ \cmidrule{3-4}
        &  & \texttt{FID}* $\downarrow$ / \texttt{IS}* $\uparrow$  &  \texttt{FID}* $\downarrow$ / \texttt{IS}* $\uparrow$   \\ \midrule
        \multirow{3}{*}{Half} &
          {\scriptsize ~TAC-GAN}    & -
          & ~23.75 / 28.86{\scriptsize  $\pm$0.29}$^ \ddagger $
          \\
            &
          {\scriptsize ~BigGAN}    & 12.75 / 21.84{\scriptsize $\pm$0.34}
          & ~22.77 / 38.05{\scriptsize  $\pm$0.79}$^\ddagger $
          \\
          & {\scriptsize ~~\textbf{\texttt{FQ}}-BigGAN}  & 
          \cellcolor{Gray}
          \textcolor{blue}{\textbf{12.62 / 21.99{\scriptsize  $\pm$0.32}}}
          & 
          \cellcolor{Gray}
          \textcolor{blue}{\textbf{~19.11 / 41.92{\scriptsize  $\pm$1.15 } }} \\
         \hline
         \multirow{2}{*}{256K} &
          {\scriptsize ~BigGAN}    & 10.55 / 25.43{\scriptsize $\pm$0.15} 
          & ~~14.88 / 63.03{\scriptsize  $\pm$1.42}$^\dagger$ 
          \\
          & {\scriptsize  ~~\textbf{\texttt{FQ}}-BigGAN}  &  \cellcolor{Gray} \textcolor{blue}{\textbf{9.67 / 25.96{\scriptsize  $\pm$0.24}}} 
          & \cellcolor{Gray} \textcolor{blue}{\textbf{ 13.77}} / 54.36{\scriptsize  $\pm$1.07} \\
    \bottomrule
    \end{tabular}
    \end{minipage}
    \caption{Comparison on ImageNet-1000 for two resolutions. Both models were trained for 256K iterations if not diverge early. The top and bottom block shows the best results within {\it half} and {\it full} of the entire training procedure, respectively. $^ \ddagger$ from~\cite{gong2019twin}, $^\dagger$ from~\citep{brock2018large}, we cannot reproduce it using their codebase, as the training diverges early.}
    \label{tab:score_imagenet}
\end{table}

\vspace{-2mm}
\subsection{BigGAN for Image Generation}
BigGAN~\citep{brock2018large} holds the state-of-the-art on the task of class-conditional image synthesis, which benefits from scaling up model size and batch size. 
Our implementation of BigGAN is based upon BigGAN-PyTorch\footnote{\scriptsize \url{https://github.com/ajbrock/BigGAN-PyTorch}}. We use the same architecture and experimental settings as BigGAN, except for adding FQ layers.  Best scores (\texttt{FID}* / \texttt{IS}*) and averaged scores (\texttt{FID} / \texttt{IS}) are reported. Standard deviations are computed over five random initializations and their average are reported from the best in each run.

\paragraph{CIFAR-10}\citep{krizhevsky2009learning} consists of 60K images at resolution $32\times32$ in 10 classes; 50K for training and 10K for testing. 500 epochs are used. The results are show in Table~\ref{tab:score_cifar10}. The FQ module improves BigGAN.  FQ-BigGAN also outperforms other strong existing GAN models, including spectral normalization (SN) GAN~\cite{miyato2018spectral}, Repulsive MMD-GAN~\cite{wang2019improving}. 

\paragraph{CIFAR-100} is a more challenging dataset with more fine-grained categories, compared to CIFAR-10.  The current best classification model achieves $91.3\%$ accuracy on this dataset~\cite{huang2019gpipe}, suggesting that the class distributions have certain support overlaps.  500 epochs are used.  We also integrate FQ into TAC-GAN~\cite{gong2019twin}, which is the current state-of-the-art on CIFAR-100. TAC-GAN improves the intra-class diversity of AC-GAN, thus particularly good at generating images with fine-grained labels. The results are show in Table~\ref{tab:score_cifar100}. Experimental settings to achieve these results are provided in~\ref{app:biggan}. The proposed FQ can improve both TAC-GAN and BigGAN. In particular, FQ significantly improves BigGAN on CIFAR-100 dataset. This is because FQ can increase intra-class diversity, as the dictionary items store a longer distribution history. We show generated image samples to illustrate the improved diversity for each class in Figure~\ref{fig:cifar100_sup} in Appendix.

\paragraph{ImageNet-1000}\citep{russakovsky2015imagenet} contains around 1.2 million images with 1000 distinct categories. We pre-process images into resolution $64\times64$ and $128\times128$ in our experiments, respectively. 100 epochs are used.
The results are show in Table~\ref{tab:score_imagenet}. It shows that FQ improves generation quality for both resolution 64 and 128.

\begin{table}[!t]
    \centering
    \begin{tabular}{c|c|c|c}
    \toprule
    Model & ImageNet & CIFAR-100 & CIFAR-10\\  \midrule
          BigGAN    & 7d16h &  12h12m & 17h37m\\
          \textbf{\texttt{FQ}}-BigGAN  & 7d19h & 12h35m & 17h50m\\
    \bottomrule
    \end{tabular}
    \caption{Training time comparison of before and after adding FQ module.  TITAN XP GPUs are used in these experiments. We train ImageNet $(64\times64$) for 256k iterations on 2 GPUs, and CIFAR-100 and CIFAR-10 on 1 GPU for 10k iterations.}
    \label{tab:time_cost}
\end{table}

\begin{figure}[t!]
	\vspace{-2mm}\centering
	\begin{tabular}{c c}
		\hspace{-3mm}
		\includegraphics[height=2.2cm]{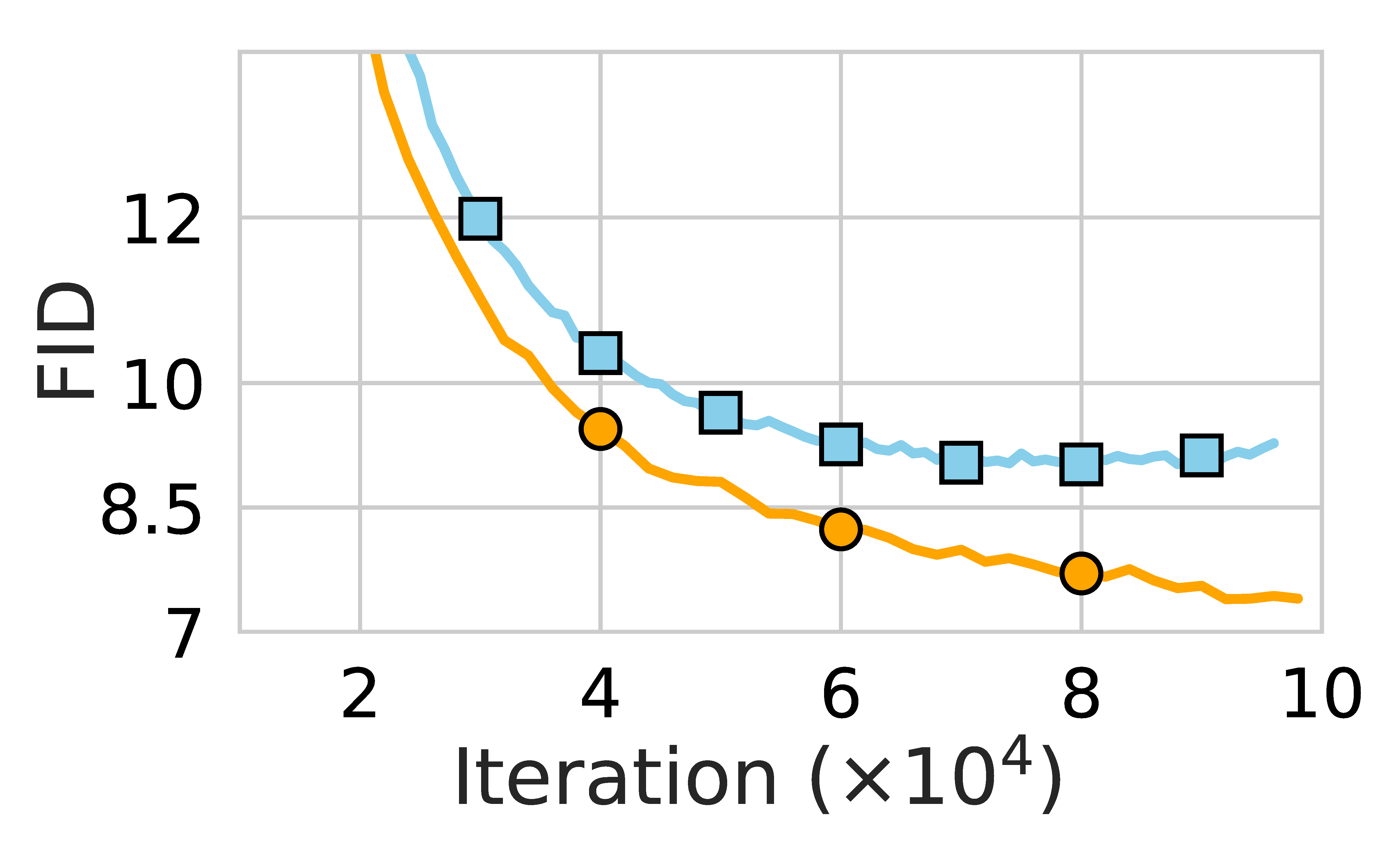}  & 
		\hspace{-6mm}
		\includegraphics[height=2.2cm]{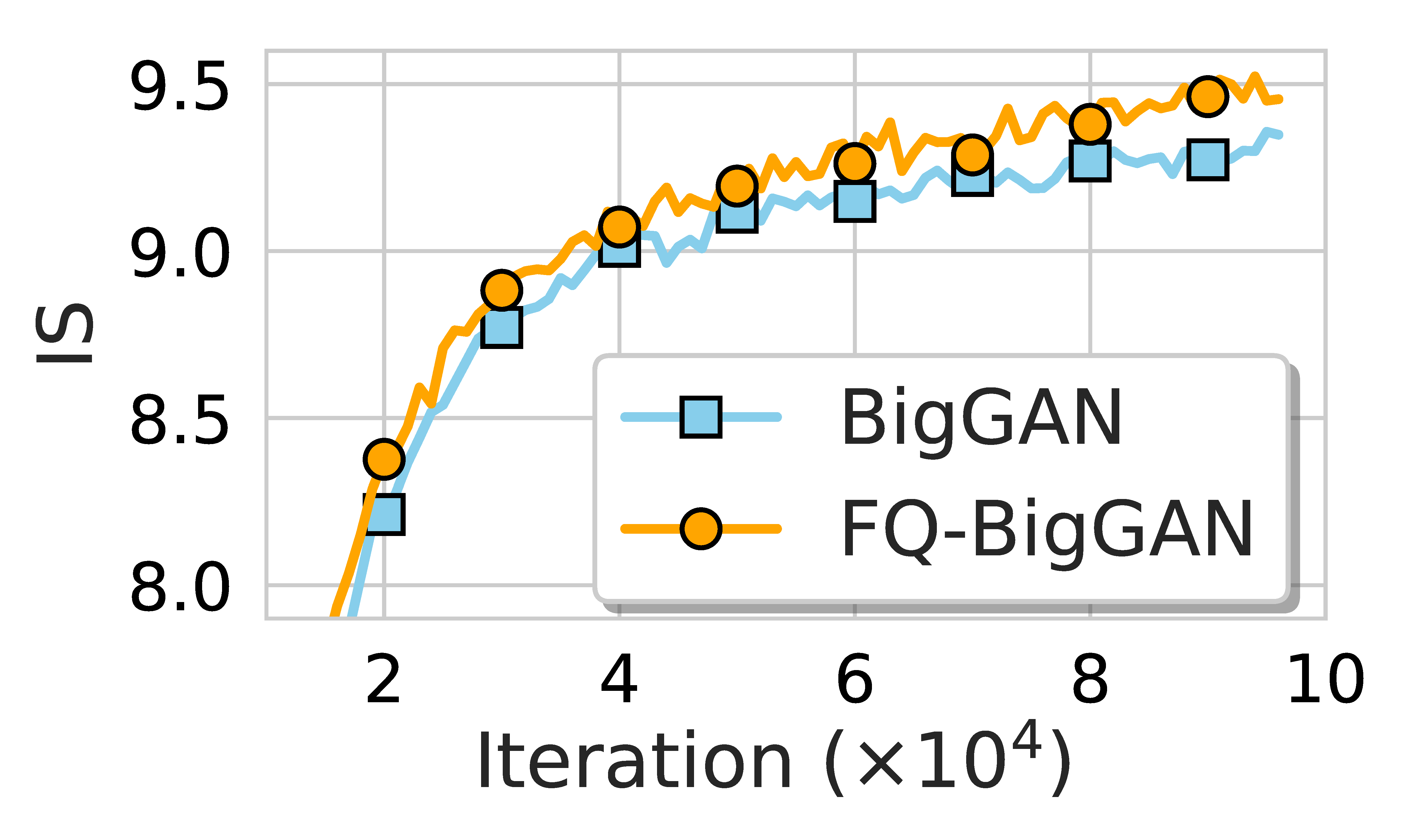} 
	\end{tabular}
	\vspace{-3mm}
	\caption{Learning curves on CIFAR-100.}
	\vspace{-3mm}
	\label{fig:learning_curve}
\end{figure}
\begin{figure}[t!]
	\vspace{-0mm}\centering
	\begin{tabular}{c}
		\hspace{-3mm}
		\includegraphics[height=2.2cm]{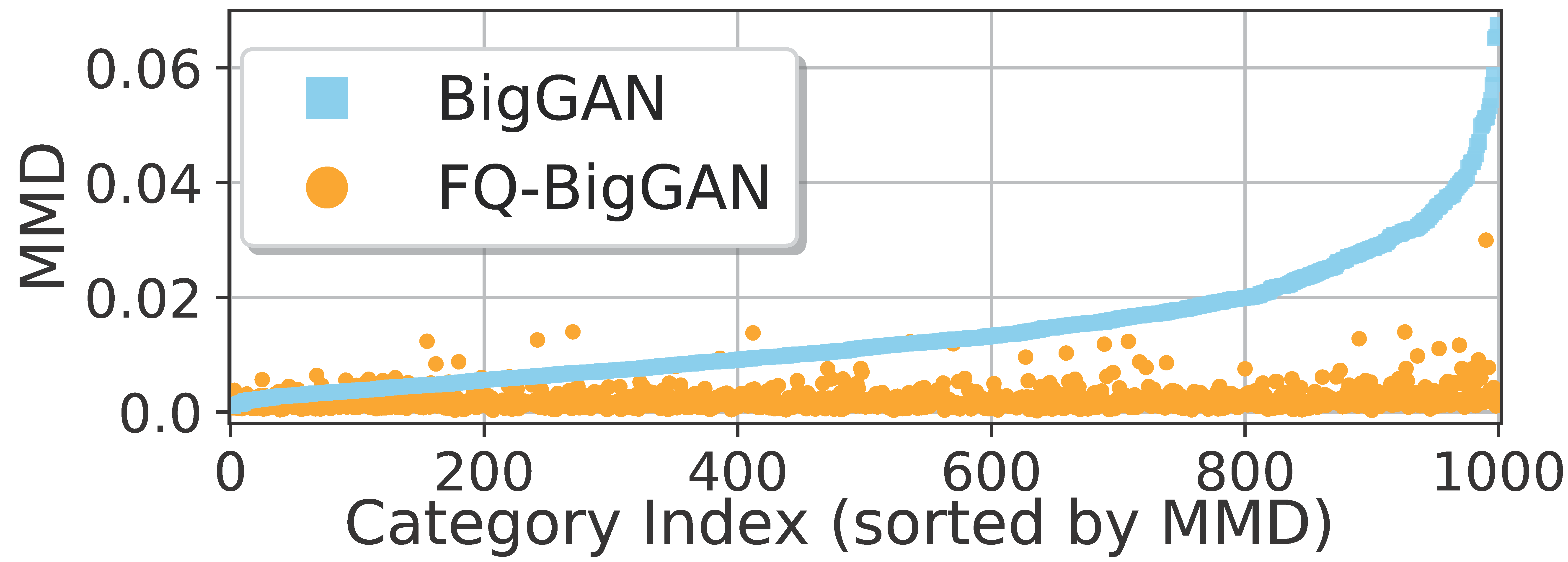} \vspace{-1mm}  \\ 
		\hspace{-3mm}
		\includegraphics[height=2.2cm]{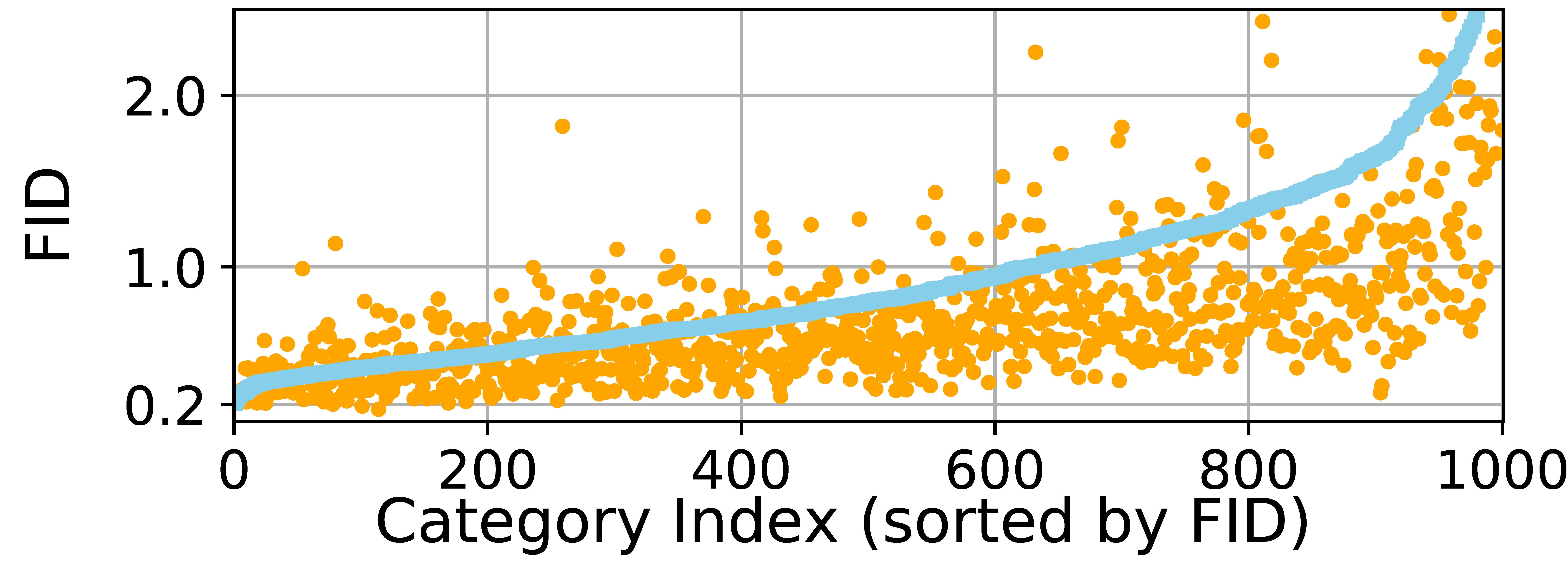} 
	\end{tabular}
	\vspace{-2mm}
	\caption{Comparison on per-class metrics for ImageNet. 
	 }
	\vspace{-2mm}
	\label{fig:imagenet_per_class}
\end{figure}

\paragraph{Computational Cost.} To evaluate the computational overhead of FQ, we compare the running time of BigGAN and our FQ-BigGAN variant on three datasets in Table~\ref{tab:time_cost}. To finish the same number of training epochs, FQ-GAN takes 1.63\%, 3.14\%, and 1.23\% more time than the original Big-GAN on ImageNet, CIFAR-100 and CIFAR-10, respectively. It means that the additional time cost of FQ is negligible. In practice, FQ-GAN converges faster, as shown in Figure~\ref{fig:learning_curve}. It may take less time to reach the same performance.

\paragraph{How does FQ improve performance?} 
We perform in-depth analysis on per-class image generation quality on ImageNet. Two metrics are studied: 
$(\RN{1})$ 
We extract quantized feature sets from discriminator for both real and fake images per class, and measure their distribution divergence using maximum mean discrepancy (MMD). Lower MMD values indicate better feature matching.
$(\RN{2})$ 
The per-class FID is also computed from the pre-trained inception network, which measures the matching quality that a generated distribution fits the target distribution in each class. Lower FID values indicate better high intra-class diversity. 
The results are shown in In Figure~\ref{fig:imagenet_per_class}. FQ yields significantly lower MMD, and lower FID by a large margin, meaning that FQ can improve feature matching, and intra-class diversity.


\begin{table}[!t]
    \centering
    \begin{tabular}{c|c|c|c|c}
    \toprule
    Resolution & $32^2$& $64^2$ & $128^2$ & $1024^2$ \\ \midrule
    StyleGAN  & 3.28 & 4.82 & 6.33 & 5.24\\
    \rowcolor{Gray} 
    \textbf{\texttt{FQ}}-StyleGAN & 
    \textcolor{blue}{\textbf{3.01}} & 
    \textcolor{blue}{\textbf{4.36}} & 
    \textcolor{blue}{\textbf{5.98}} &
    \textcolor{blue}{\textbf{4.89}}\\
    \bottomrule
    \end{tabular}
     \vspace{-2mm}
    \caption{StyleGAN: Best FID-50k scores in FFHQ at different resolutions.}
    \label{tab:ffhq}
    \vspace{-2mm}
\end{table}

\begin{figure*}[t!]
\begin{minipage}[b]{0.7\textwidth} 
\centering
 \includegraphics[height=4.2cm]{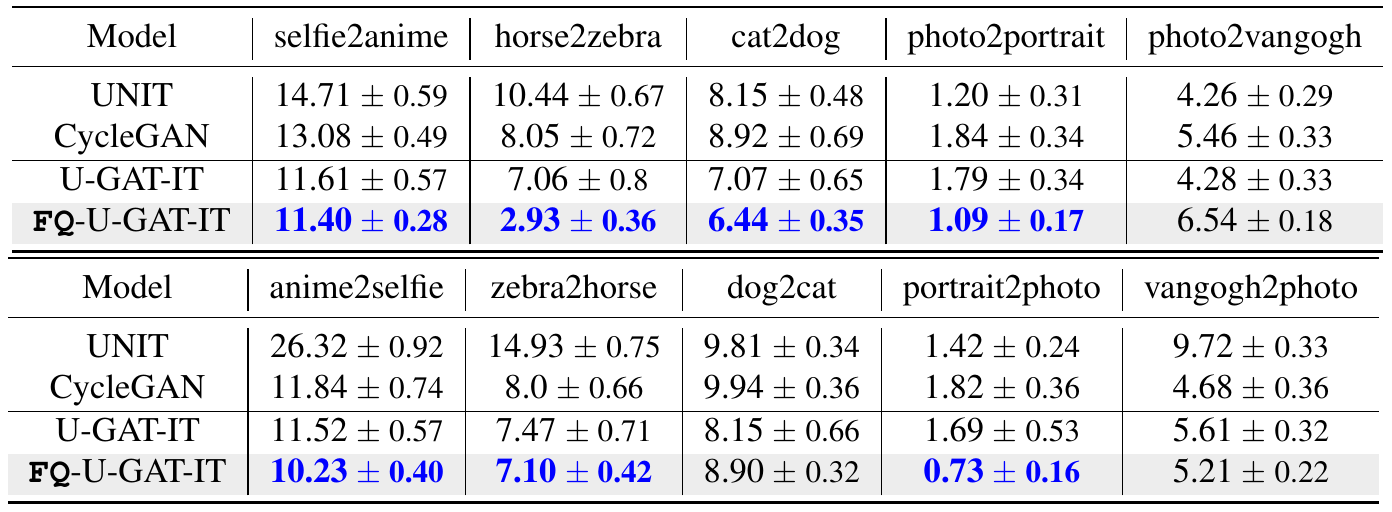}
    \vspace{-2mm}
    \captionof{table}{KID $\times 100 $ for different image translation datasets. All numbers except for our FQ variant are from~\cite{kim2019u}. }
    \label{tab:i2i_kid}
\end{minipage}
\hfill 
\begin{minipage}[b]{0.25\textwidth} 
\centering
 \includegraphics[height=2.5cm]{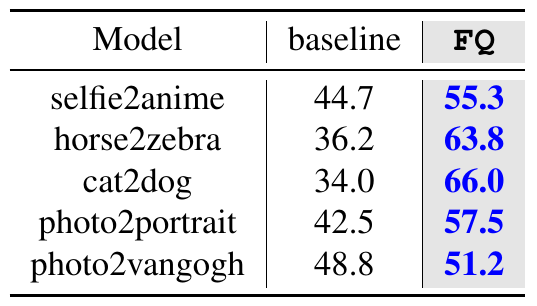}
    \vspace{-3mm}
    \captionof{table}{User perceptual study on translated image preference (in percentage) between U-GAT-IT and its FQ variant using AMT.
    \\
    \hspace{20mm}}
    \label{tab:i2i_userstudy}
\end{minipage}
\end{figure*}

\begin{figure*}[!t]
    \centering
    \includegraphics[width=0.95\textwidth]{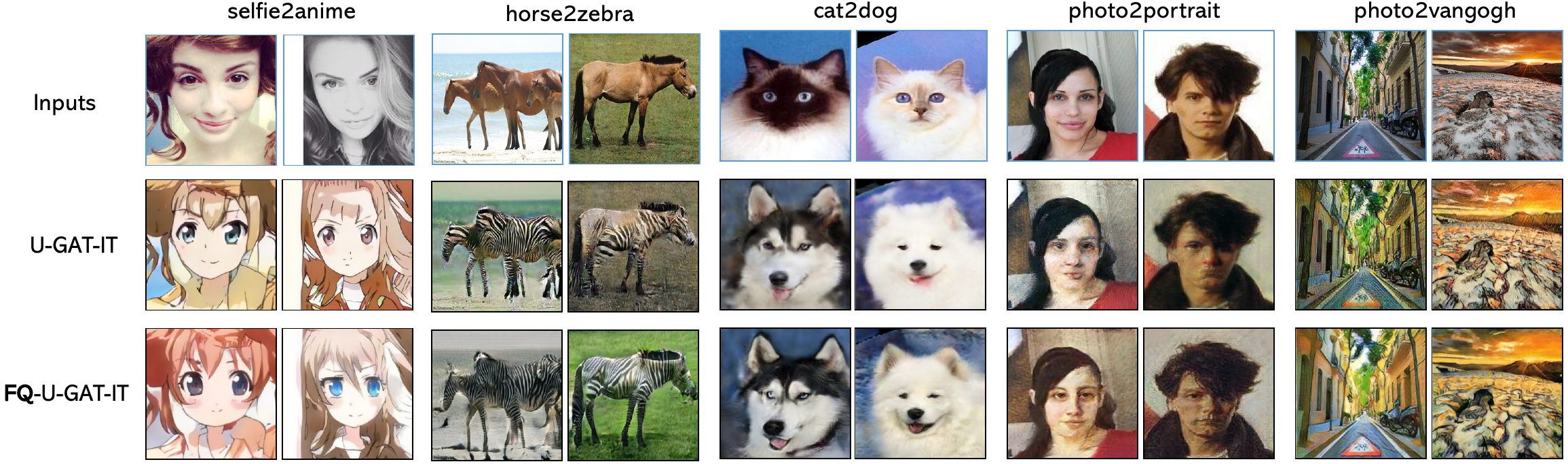}
    \caption{Qualitative comparison. The 1st, 2nd and 3rd shows source and the translated images using U-GAT-IT and FQ, respectively. }
    \label{fig:i2i_samples}
\end{figure*}

\subsection{StyleGAN for Face Synthesis}
StyleGAN yields state-of-the-art results in unconditional generative image modeling. 
StyleGAN~\citep{karras2019style} is a new variant of the Progressive GAN \citep{karras2018progressive}, the main difference is the introduction of a latent mapping network in the generator architecture. 
The very recent version, StyleGAN2~\cite{karras2019analyzing}, simplifies the progressive architecture, and uses a suite of techniques to improve the performance.
We apply our FQ to the StyleGAN and StyleGAN2, based on the TensorFlow codes of StyleGAN \footnote{\scriptsize \url{https://github.com/NVlabs/stylegan}}
and StyleGAN2\footnote{\scriptsize \url{https://github.com/NVlabs/stylegan2}}. 
The Flickr-Faces-HQ (FFHQ) dataset \citep{karras2019style} is used. It consists of 70k high-quality images ($1024\times1024$), which endows more variations than the previously widely used CelebA-HQ dataset in terms of accessories, age, ethnicity and image background  \citep{karras2019style}. Each model was trained using 25M images by default.
For StyleGAN, we consider four resolutions at $32^2, 64^2$, $128^2$ and $1024^2$. The progressive training starts from resolution $8^2$ in experiments of resolution $32^2-128^2$ whereas the initial resolution is $512^2$ in the experiment on $1024^2$. The results are shown in Table~\ref{tab:ffhq}. The FQ variant improves StyleGAN on all four resolutions. For StyleGAN2, we deploy the model under \textit{config-e} ~\cite{karras2019analyzing} and use the full resolution FFHQ. The best FID score of \textbf{\texttt{FQ}}-StyleGAN2 is \textcolor{blue}{{\bf 3.19}} which surpasses the reported score 3.31 of StyleGAN2. High-fidelity generated faces from the two models are given in Appendix.

\subsection{Unsupervised Image-to-Image Translation}
The task of unsupervised image translation is becoming increasingly popular, inspired by recent advances in GANs. U-GAT-IT~\cite{kim2019u} is the latest state-of-the-art. 
We validate our FQ using their official TensorFlow codebase\footnote{\scriptsize \url{https://github.com/taki0112/UGATIT}}. 
Five unpaired image datasets are used for evaluation, including selfie2anime \citep{kim2019u}, cat2dog, photo2portrait~\citep{lee2018diverse}, horse2zebra and vangogh2photo \citep{zhu2017unpaired}.  
All images are resized to $256 \times 256$ resolution. Details are given in the Appendix.

We also compare with several known image translation models, including CycleGAN \citep{zhu2017unpaired} and UNIT \citep{huang2018multimodal}, which show better performance than MUNIT  \citep{liu2017unsupervised}, DRIT \citep{lee2018diverse} in~\citep{kim2019u}. Each model is trained for 100 epochs, and we report the best KID values in Table~\ref{tab:i2i_kid}. FQ improves U-GAT-IT on most datasets, and achieves new state-of-the-art for image translation. 
We have also conducted human evaluation on Amazon Mechanical Turk (AMT). Each testing image is judged by 3 users, who are asked to select the best translated image to target domain. We inform to the participants the name of target domain, along with six example images of target domain as visual illustration. An example of user interface is show in Figure~\ref{fig:amt_interface} in Appendix.
Table~\ref{tab:i2i_userstudy} shows the overall percentage that users prefer a particular model. The proposed FQ achieves higher score in human perceptual study (except for comparable results on photo2vangogh), compared to its baseline method. 
Qualitative comparison in Figure~\ref{fig:i2i_samples} shows that FQ can produce sharper image regions, this is because the dictionary items that FQ utilizes to construct features are from recent history. More examples on translated images are in Appendix.


\section{Conclusion}
In this paper, we propose Feature Quantization Generative Adversarial Networks (FQ-GANs), which incorporate a feature quantization module into the discriminator learning of the GAN framework. 
The FQ module is effective in performing implicit feature matching for large datasets. FQ can be easily used in training many existing GAN models, and improve their performance. It yields improved performance on three canonical tasks, including BigGAN for image generation on ImageNet,  StyleGAN and StyleGAN2 for face generation on FFHQ, and unsupervised image-to-image translation. FQ-GAN sets new state-of-the-art performance on most datasets.

\section*{Acknowledgements}
The authors gratefully acknowledge Yanwu Xu for preparing the TAC-GAN codebase, and Yulai Cong for proofreading the draft. We are also grateful to the entire Philly Team inside Miscrosoft for providing our computing platform.

\bibliography{references.bib}
\bibliographystyle{icml2020}


\twocolumn[
\icmltitle{\Large Appendix:
 Feature Quantization Improves GAN Training}
]

\appendix


\section{BigGAN}
\subsection{More results on Ablation Study}
\begin{figure*}[t!]
	\vspace{-0mm}\centering
	\begin{tabular}{c c}
		\includegraphics[width=0.45\textwidth]{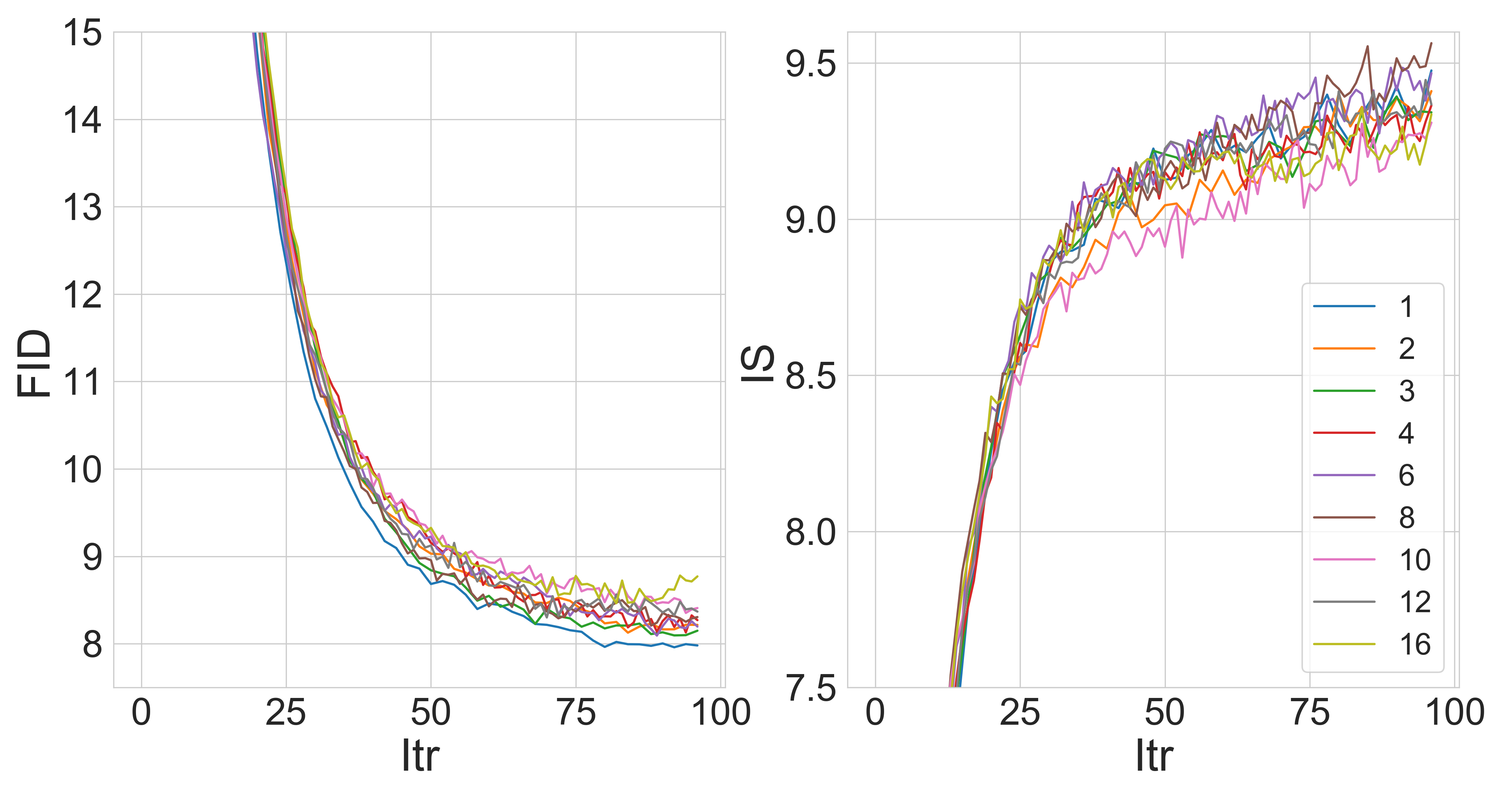}  & 
		\includegraphics[width=0.45\textwidth]{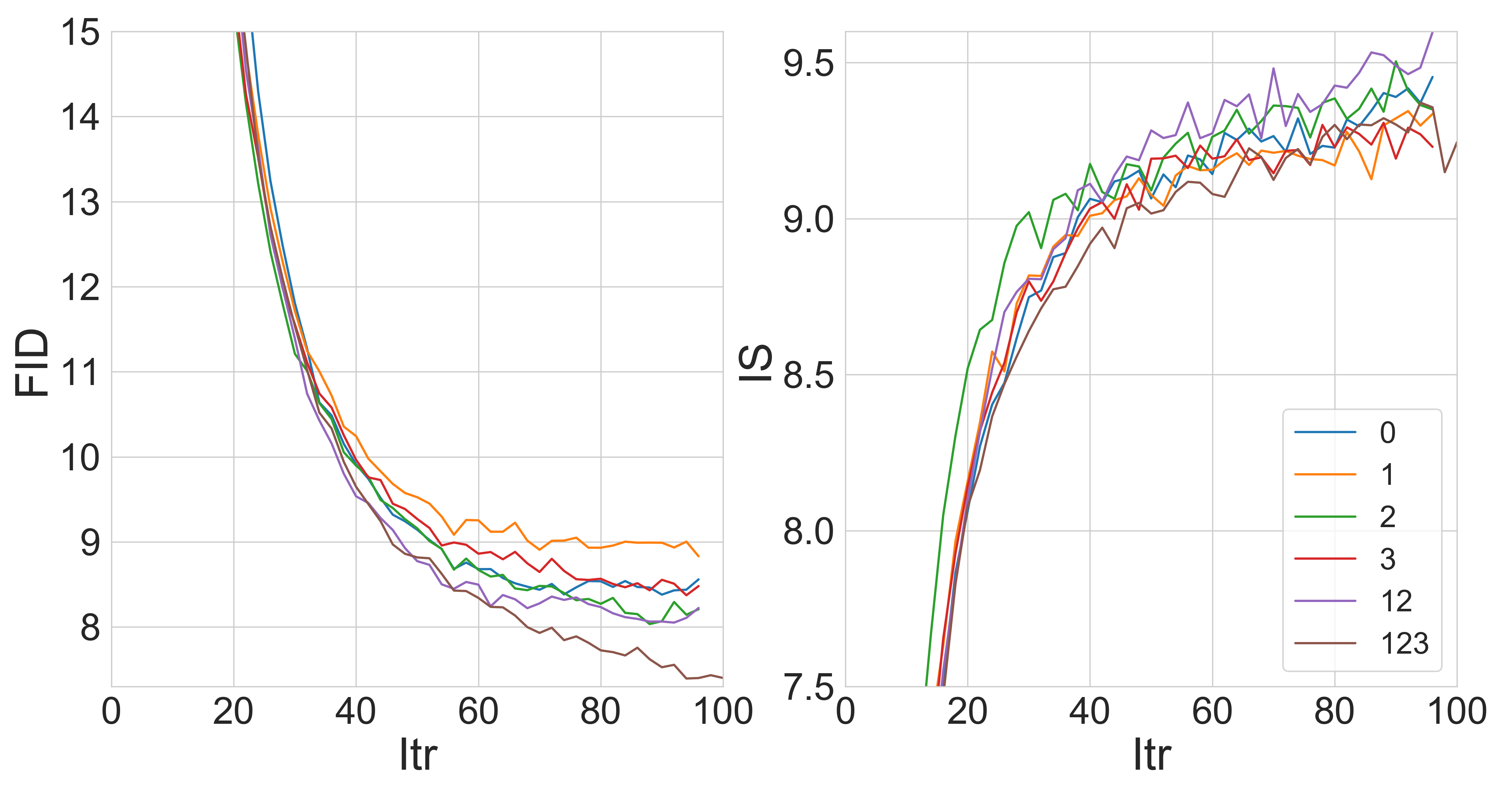}  \\
		(a) Dictionary size $P$\vspace{0mm} & 
		(b) FQ layer position \hspace{-0mm} \\
		\includegraphics[width=0.45\textwidth]{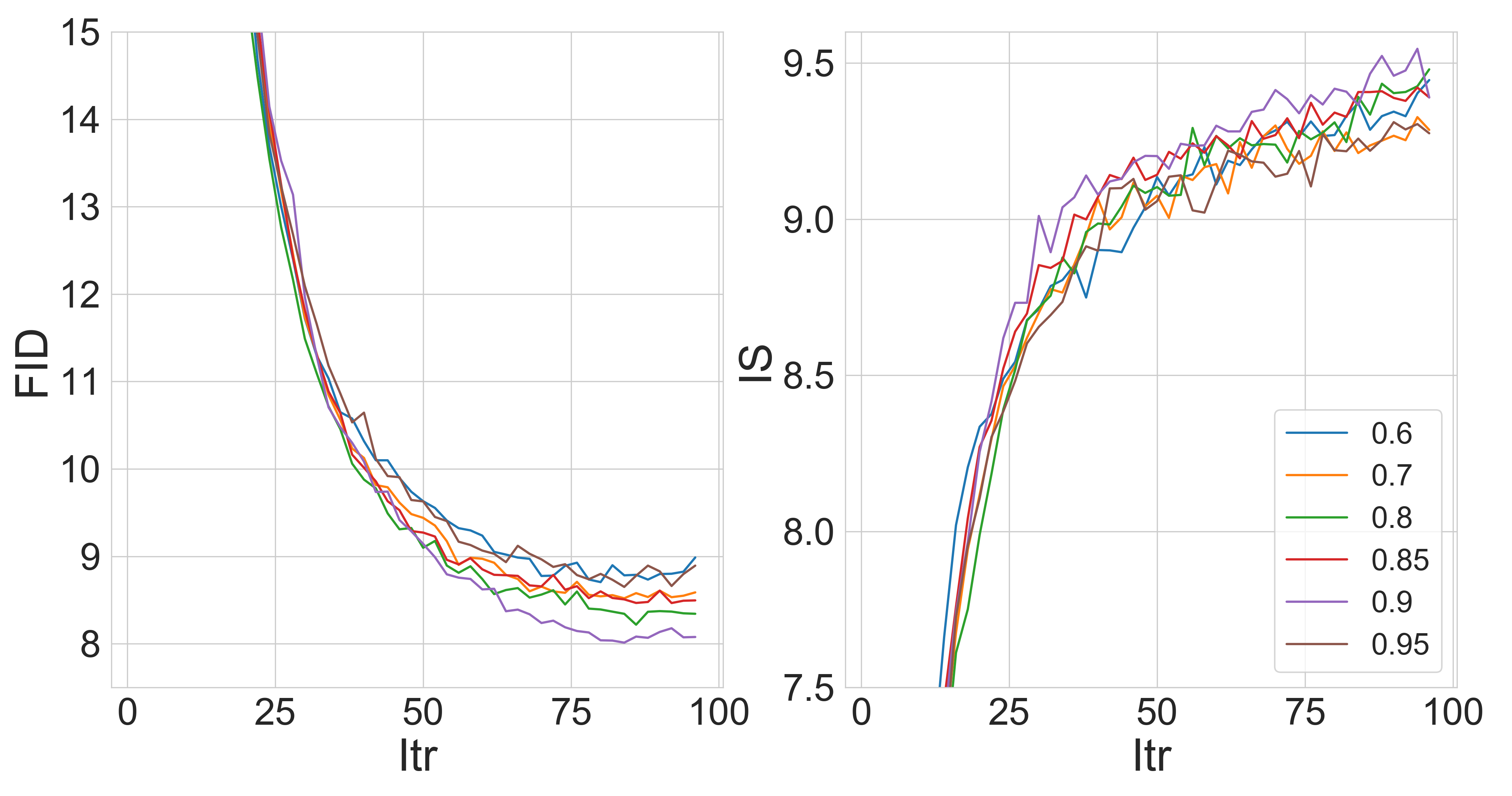}  &
		\includegraphics[width=0.45\textwidth]{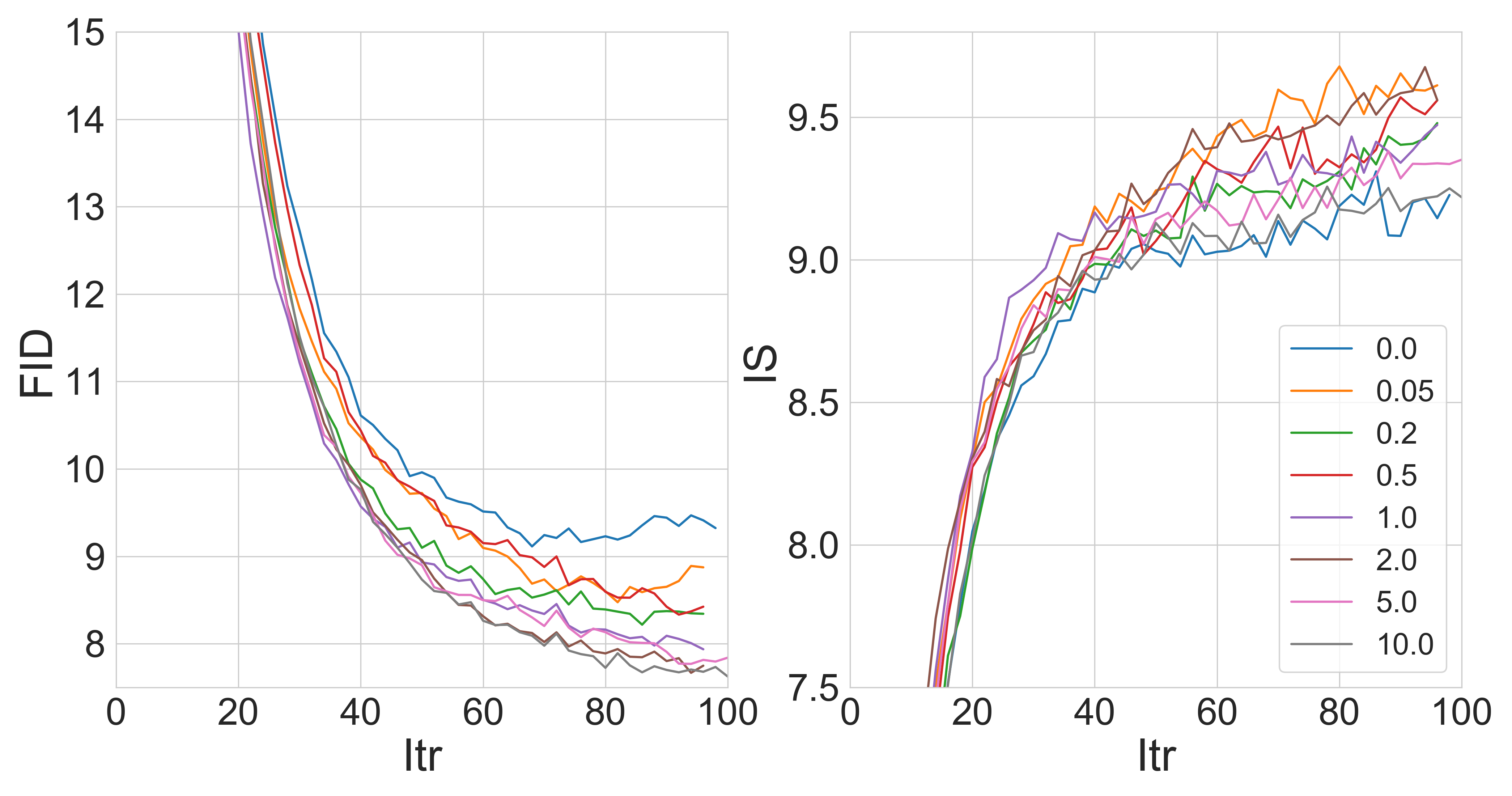} \\
		(c) $\lambda$ \hspace{0mm} & 
		(d) $\alpha$ \hspace{-0mm} \\ 
	\end{tabular}
	\vspace{-0mm}
	\caption{Ablation studies on the impact of hyper-parameters. The image generation quality is measured with FID $\downarrow$ and IS $\uparrow$. (a) Dictionary size $K=2^P$. (b) The positions to apply FQ. (c) The decay hyper-parameter $\lambda$ in momentum-based dictionary update. (d) The weight $\alpha$ to incorporate FQ,.
	 }
	\vspace{-0mm}
	\label{fig:ablation_sup}
\end{figure*}

In Figure \ref{fig:ablation_sup}, we provide the detailed learning curves under different FQ settings on CIFAR100.

\subsection{Experiment setup}\label{app:biggan}
\begin{itemize}
    \item CIFAR-10 and CIFAR-100 ($32\times32$ ): $bs=64, ch=64$. The architecture is given in Table \ref{tab:arch_cifar}. Parameters are set as: $bs=64, G\_lr=2e^{-4}, D\_lr=2e^{-4}, D\_step=4, G\_step=1$. To get the best results shown in Table~\ref{tab:score_cifar100}, we set $P=10, \lambda=0.9, \alpha=1.0$ of FQ being added at the layers [0, 1, 2, 3]. 
    \item ImageNet ($64\times64$): $bs=512, ch=64$. The architecture is the same as that in Imagenet ($128\times128$) when you omit the bottom downsample ResBlock in the discriminator and the top upsample ResBlock in the generator, as shown in Table. \ref{tab:arch_imagenet}. Parameters are set as: $bs=512, G\_lr=e^{-4}, D\_lr=4e^{-4}, D\_step=1, G\_step=1$ with self-attention at resolution $32 \times 32$. $P=10, \lambda=0.7, \alpha=1.0$ of FQ.
    \item Imagenet ($128\times128$): The architecture is given in Table. \ref{tab:arch_imagenet}. Due to limited hardware resources, compared with the full-version BigGAN, we did the following modification: $bs=2048 \xrightarrow{} bs=1024, ch=96 \xrightarrow{} ch=64$. $P=10, \lambda=0.8, \alpha=10.0$ of FQ.s

\end{itemize}

\subsection{Generated image samples}
\begin{figure*}[t!]
	\vspace{-0mm}\centering
	\begin{tabular}{c c c c}
		\includegraphics[width=0.22\textwidth]{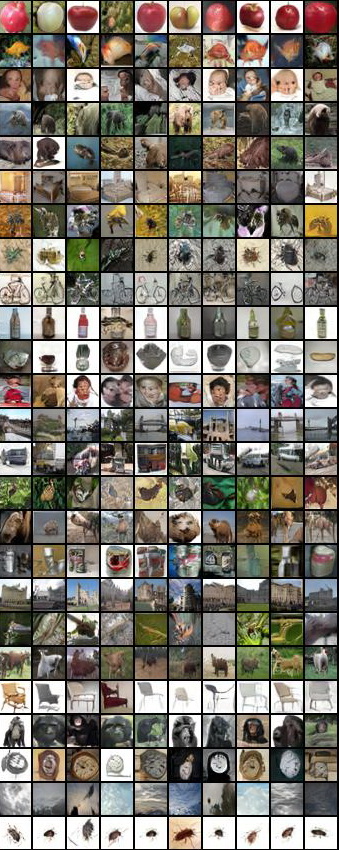}  & 
		\includegraphics[width=0.22\textwidth]{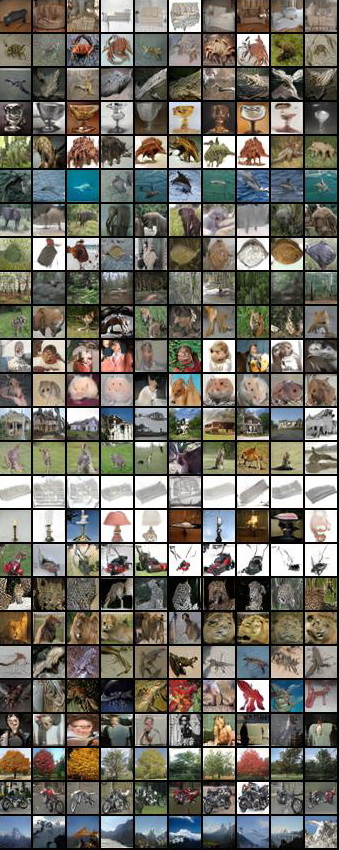}  &
		\includegraphics[width=0.22\textwidth]{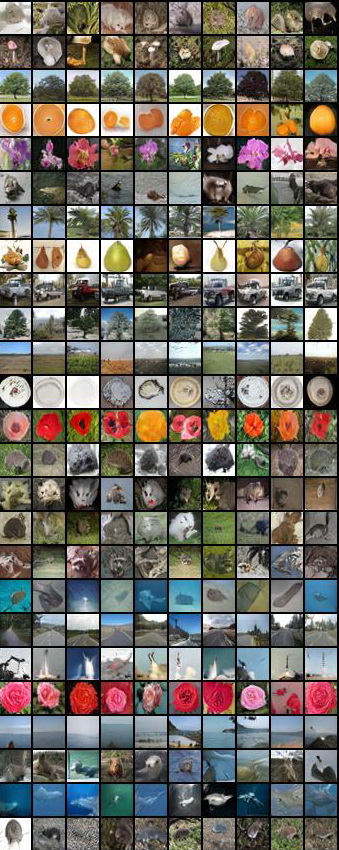}  &
		\includegraphics[width=0.22\textwidth]{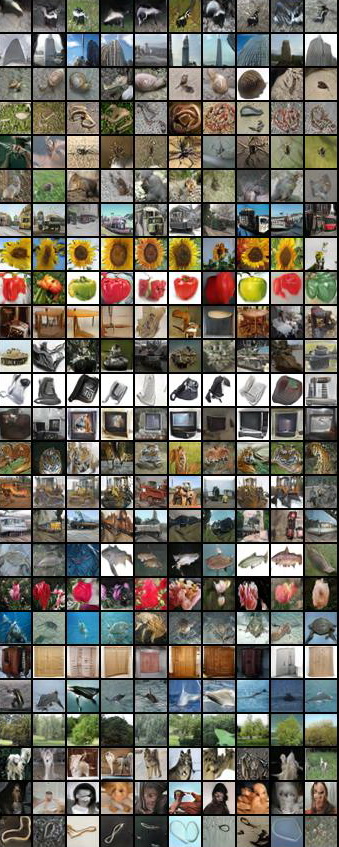}  \\ \bottomrule
		\addlinespace[1ex] \includegraphics[width=0.22\textwidth]{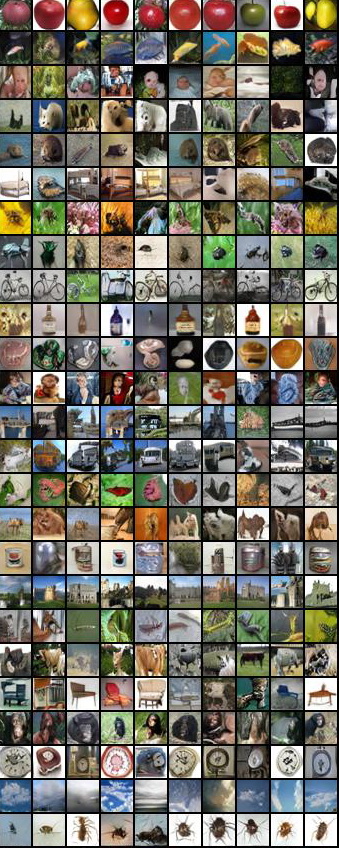}  &
		\includegraphics[width=0.22\textwidth]{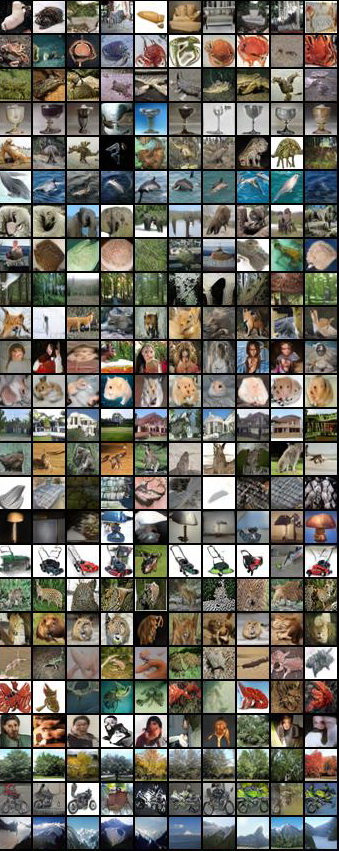}  &
		\includegraphics[width=0.22\textwidth]{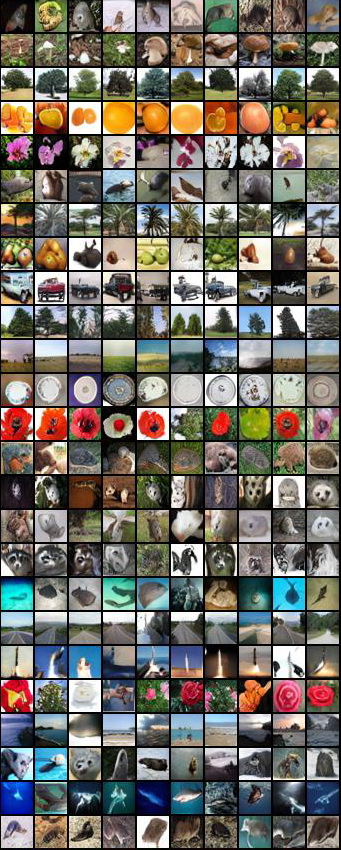}  &
		\includegraphics[width=0.22\textwidth]{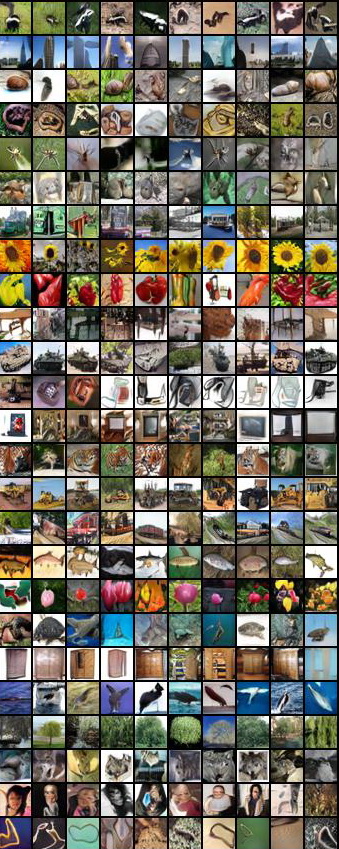} \\
	\end{tabular}
	\vspace{-0mm}
	\caption{Conditionally generated samples (under lowest FID) of BigGAN and FQ-BigGAN on CIFAR-100. (\textbf{Top} BigGAN, \textbf{Bottom} FQ-BigGAN). FQ-BigGAN obviously surpasses the BigGAN in sample diversity and fidelity.}
	\vspace{-0mm}
	\label{fig:cifar100_sup}
\end{figure*}

\begin{figure*}[!t]
	\vspace{-0mm}\centering
	\begin{tabular}{c | c}
		\includegraphics[width=0.45\textwidth]{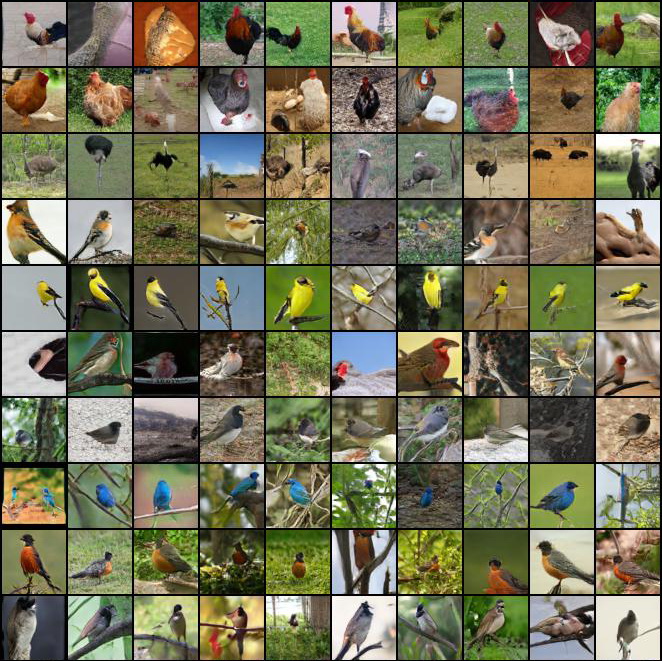}  & 
		\includegraphics[width=0.45\textwidth]{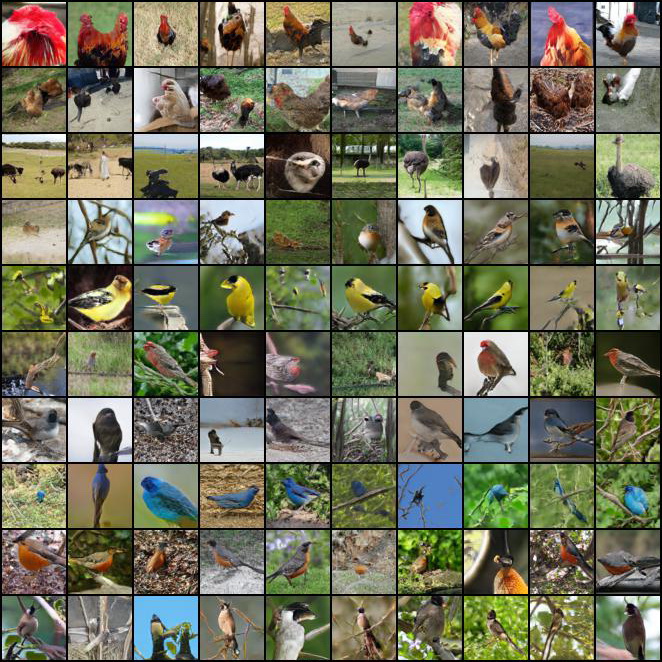}  \\
		\includegraphics[width=0.45\textwidth]{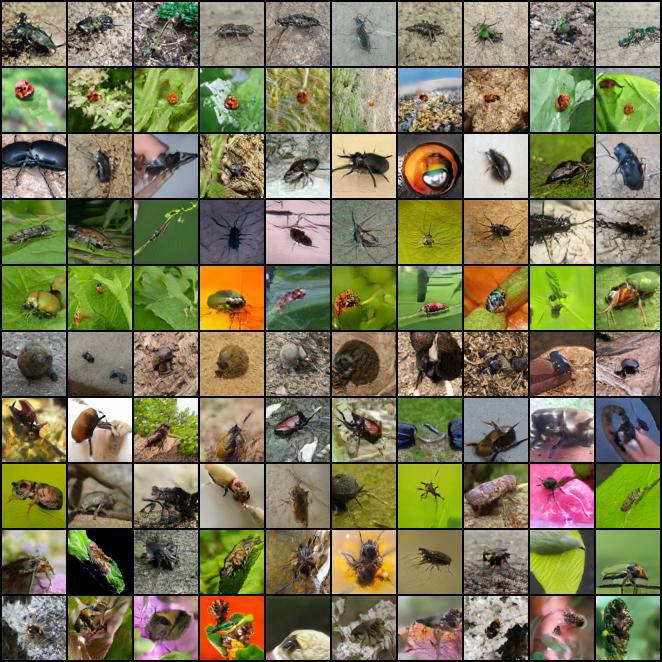}  &
		\includegraphics[width=0.45\textwidth]{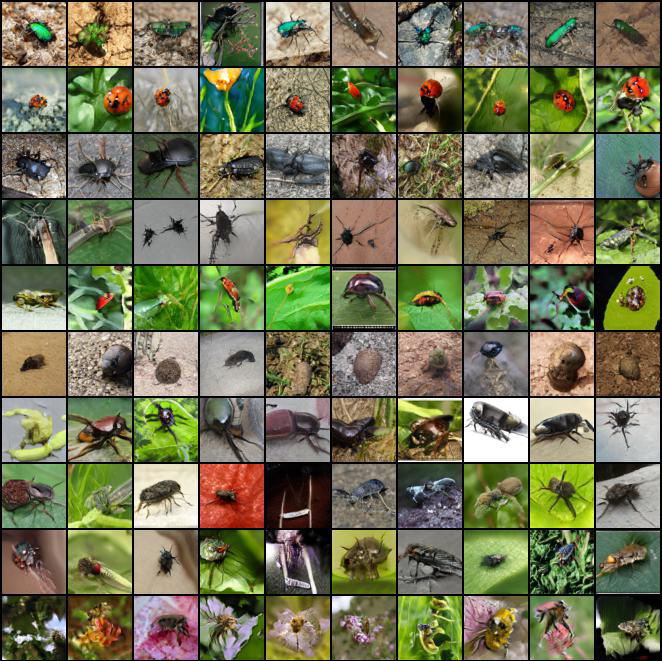} \\
		(a) BigGAN \hspace{0mm} & 
		(b) FQ-BigGAN \hspace{-0mm} \\ 
	\end{tabular}
	\vspace{-0mm}
	\caption{Conditionally generated samples of BigGAN and FQ-BigGAN on ImageNet. FQ-BigGAN can generate more diverse and accurate samples than BigGAN.}
	\vspace{-0mm}
	\label{fig:imagenet_sup}
\end{figure*}

\begin{figure*}[!t]
	\vspace{-0mm}\centering
	\begin{tabular}{c}
		\includegraphics[width=0.95\textwidth]{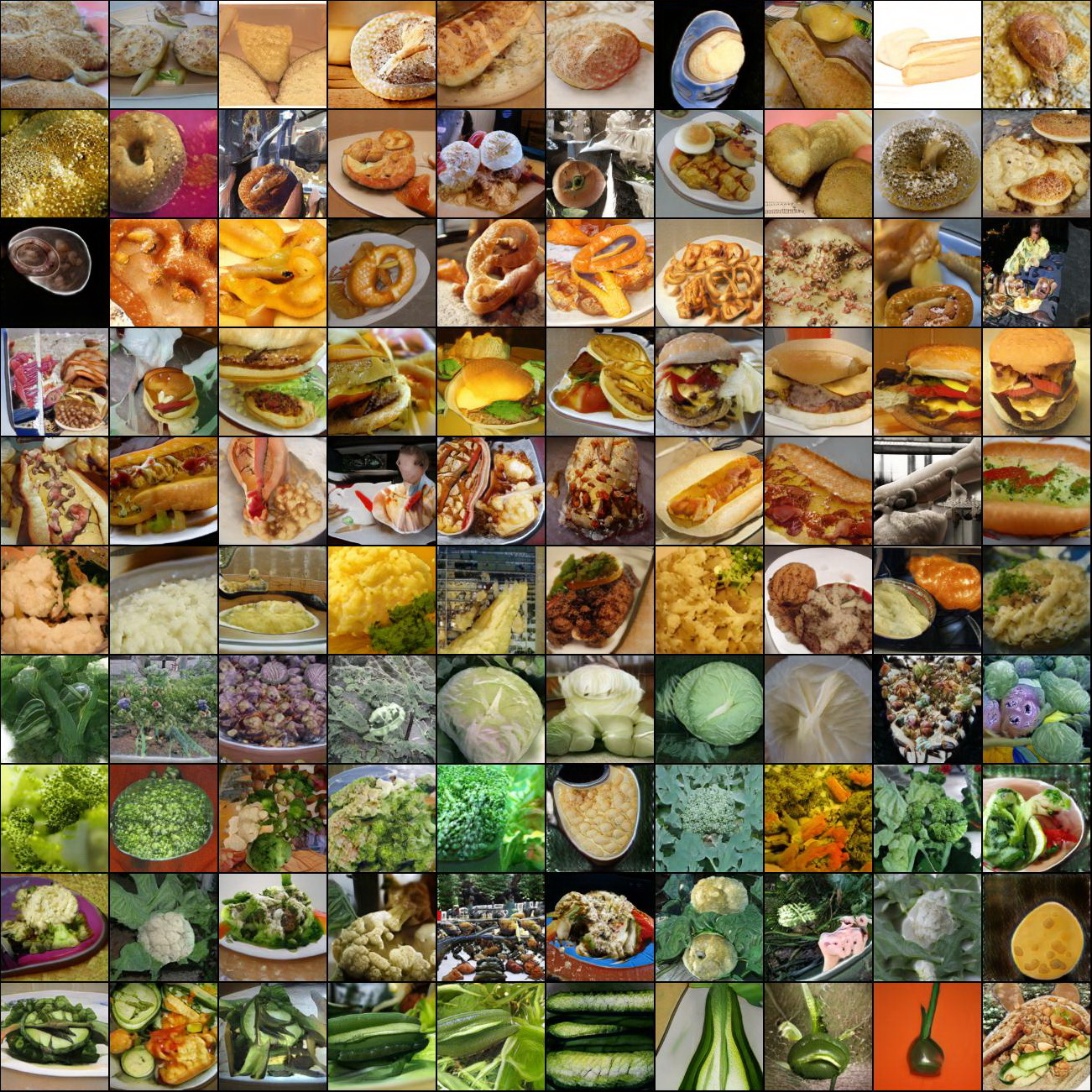}  \\ 
	\end{tabular}v
	\vspace{-0mm}
	\caption{More conditionally generated samples of FQ-BigGAN on ImageNet.}
	\vspace{-0mm}
	\label{fig:imagenet_sup_128_burger}
\end{figure*}

\begin{figure*}[!t]
	\vspace{-0mm}\centering
	\begin{tabular}{c}
		\includegraphics[width=0.95\textwidth]{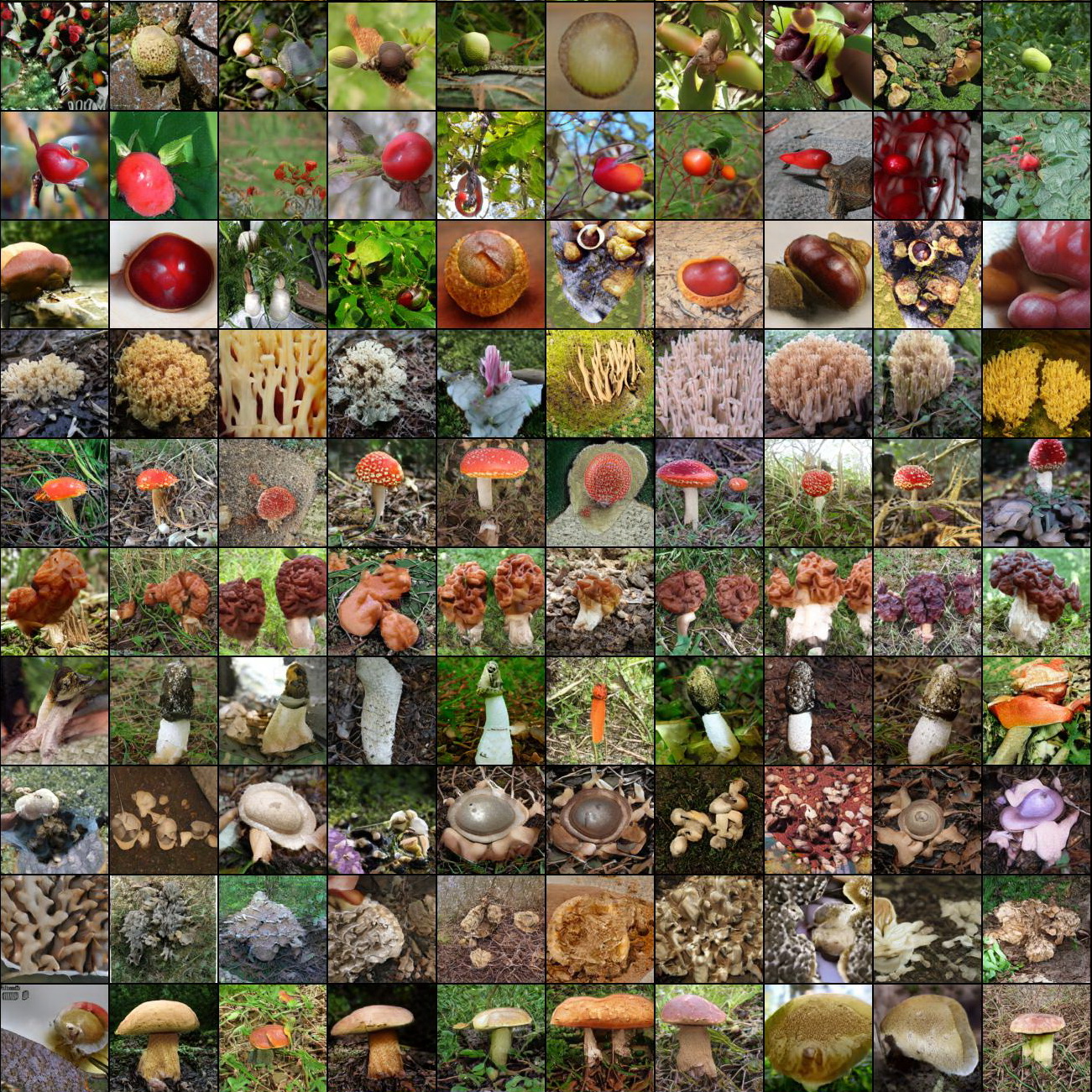}  \\
	\end{tabular}
	\vspace{-0mm}
	\caption{More conditionally generated samples of FQ-BigGAN on ImageNet.}
	\vspace{-0mm}
	\label{fig:imagenet_sup_128_mushroom}
\end{figure*}

We show the generated images for CIFAR-100  in Figure~\ref{fig:cifar100_sup}, and ImageNet in Figure~\ref{fig:imagenet_sup}. More high-fidelity results are shown in Figure~\ref{fig:imagenet_sup_128_burger} and Figure~\ref{fig:imagenet_sup_128_mushroom}.

\section{StyleGAN}
\begin{figure*}[!t]
	\vspace{-0mm}\centering
	\begin{tabular}{c}
		\includegraphics[width=0.9\textwidth]{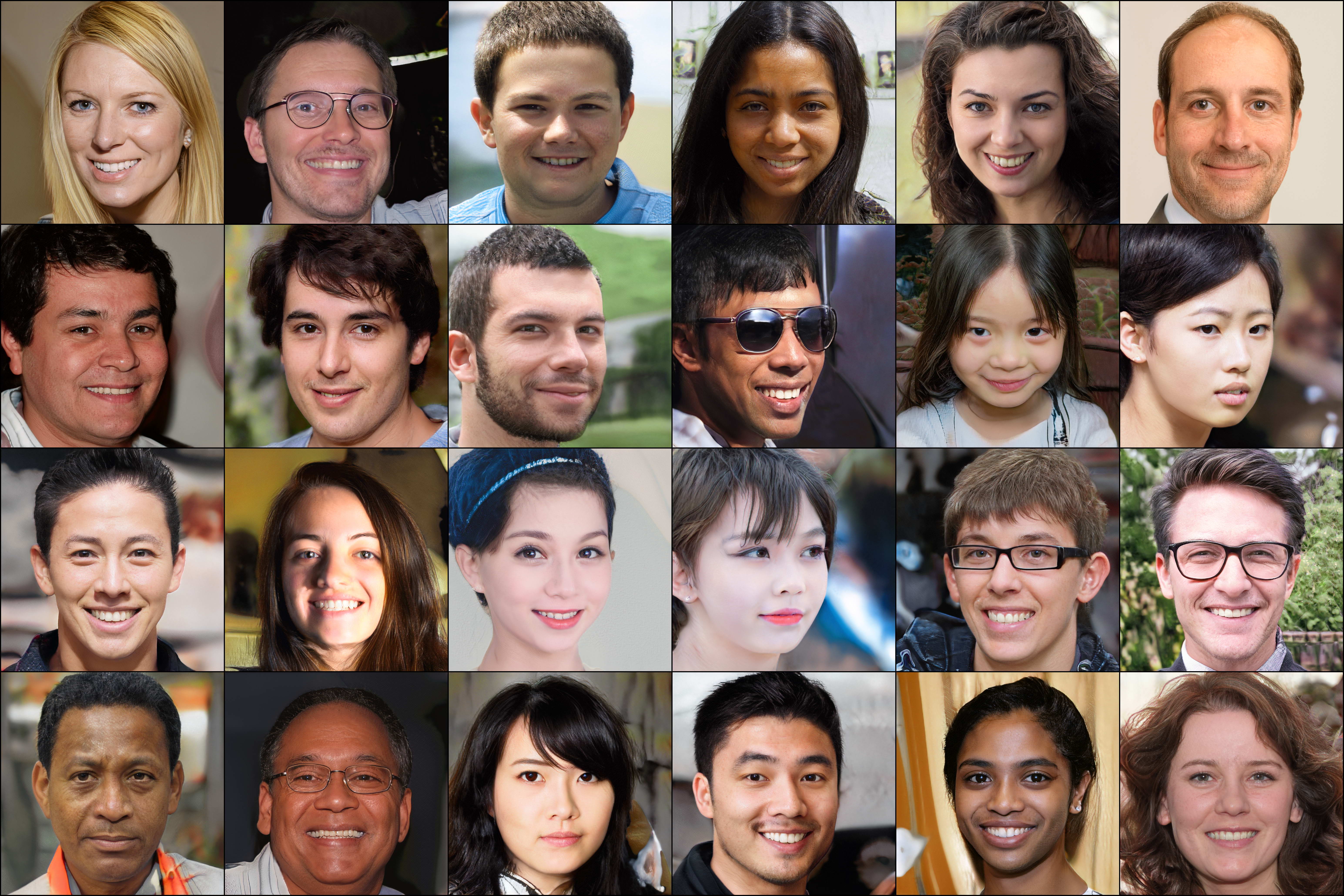}  \\
	\end{tabular}
	\vspace{-0mm}
	\caption{Images generated with \textbf{\texttt{FQ}}-StyleGAN on FFHQ-$1024^2$.}
	\vspace{-0mm}
	\label{fig:stylgean_sheet}
\end{figure*}

\begin{figure*}[!t]
	\vspace{-0mm}\centering
	\begin{tabular}{c}
		\includegraphics[width=0.9\textwidth]{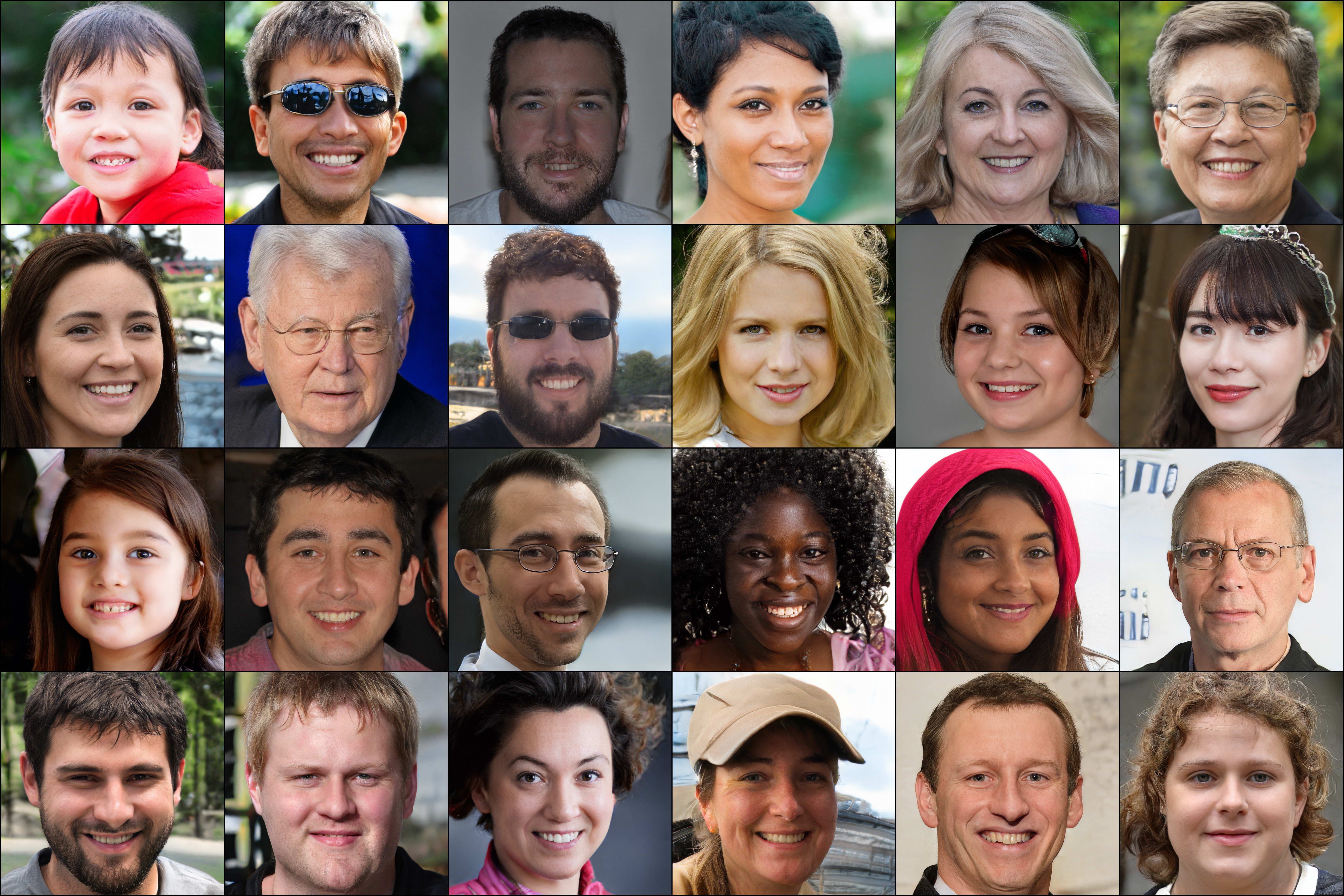}  \\
	\end{tabular}
	\vspace{-0mm}
	\caption{Images generated with \textbf{\texttt{FQ}}-StyleGAN2 on FFHQ-$1024^2$.}
	\vspace{-0mm}
	\label{fig:stylgean2_sheet}
\end{figure*}

The official discriminator architectures used in StyleGAN and StylgeGAN2 are shwon in Table~\ref{tab:stylegan_arch}. To apply the FQ techniqure, we did the following minimal modifications:
\paragraph{\textbf{\texttt{FQ}}-StyleGAN} In experiments on resolution $32^2-128^2$, we put the FQ layer just after Blocks-8 and $P=10, \lambda=0.8, \alpha=1.0$ of FQ. In experiments on resolution $1024^2$, the FQ layers were put in Blocks-(16, 32) and $P=7, \lambda=0.9, \alpha=0.25$. Randomly selected samples are shown in Figure~\ref{fig:stylgean_sheet}.
\paragraph{\textbf{\texttt{FQ}}-StylgeGAN2} We put the FQ layer in Blocks-(16, 32) and $P=7, \lambda=0.8, \alpha=0.25$ of FQ. Randomly selected samples are shown in Figure~\ref{fig:stylgean2_sheet}.

\section{U-GAT-IT}
\subsection{Dataset}

\paragraph{selfie2anime} It is first introduced in \citep{kim2019u}. The selfie and anime datasets each contains 3400 training images and 100 testing images.
\paragraph{horse2zebra and photo2vangogh} These datasets are used in \citep{zhu2017unpaired}. The training dataset size of each class: 1,067 (horse), 1,334 (zebra), 6,287 (photo), and 400 (vangogh). The test datasets consist of 120 (horse), 140 (zebra), 751 (photo), and 400 (vangogh). 
\paragraph{cat2dog and photo2portrait} These datasets are used in DRIT \citep{lee2018diverse}. The numbers of data for each class are 871 (cat), 1,364 (dog), 6,452 (photo), and 1,811 (vangogh). Follow \citep{kim2019u}, we use 120 (horse), 140 (zebra), 751 (photo), and 400 (vangogh) randomly selected images as test data, respectively.

\subsection{Architecture}
In brief, the U-GAT-IT consists of a generator, a global discriminator and a local discriminator for source to target domain translation and vice versa. We only inject our FQ into the global discriminator and keep other parts unchanged. Training settings are the same as U-GAT-IT. The modified global discriminator architecture is shown in Table \ref{tab:arch_ugatit} and $P=8, \lambda=0.8, \alpha=1.0$ of FQ.

\subsection{Additional results}

\begin{figure*}[!htbp]
    \centering
    \includegraphics[width=0.98\textwidth]{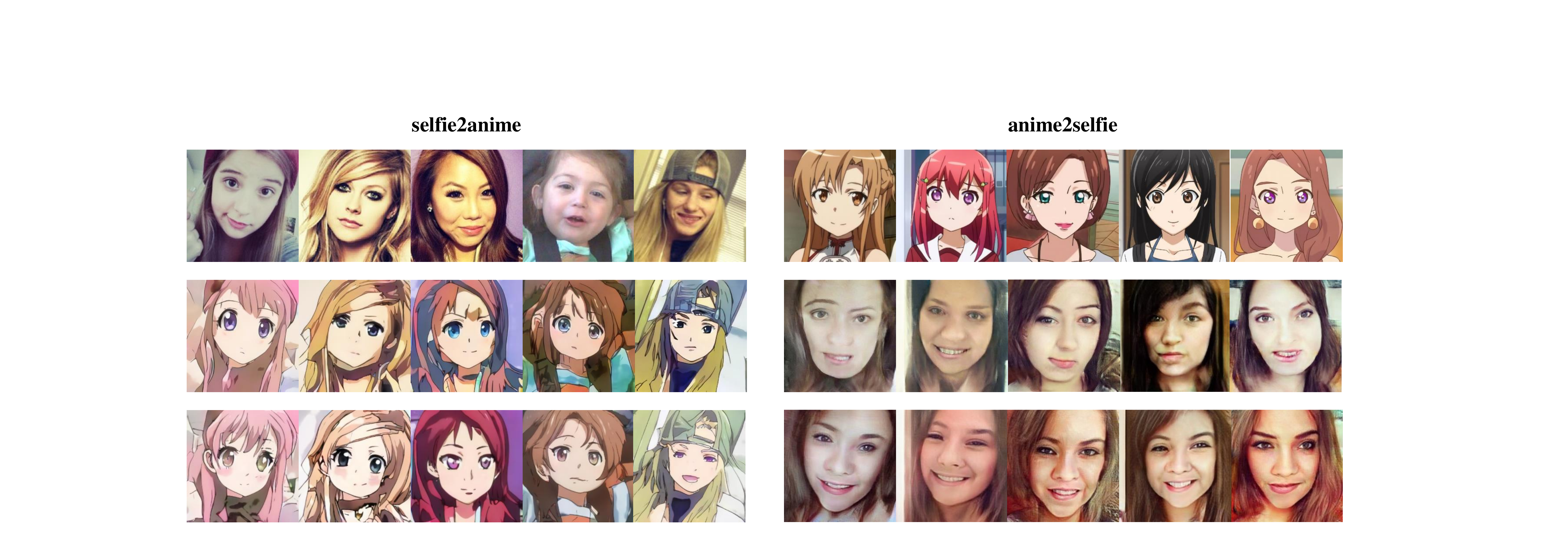}
    \caption{Visual comparisons on selfie2anime and anime2selfie. \textbf{First row}: input images. \textbf{Second row}: images generate by U-GAT-IT. \textbf{Third row}: images generated by FQ-U-GAT-IT.}
    \label{fig:i2i_anime}
\end{figure*}

\begin{figure*}[!htbp]
    \centering
    \includegraphics[width=0.98\textwidth]{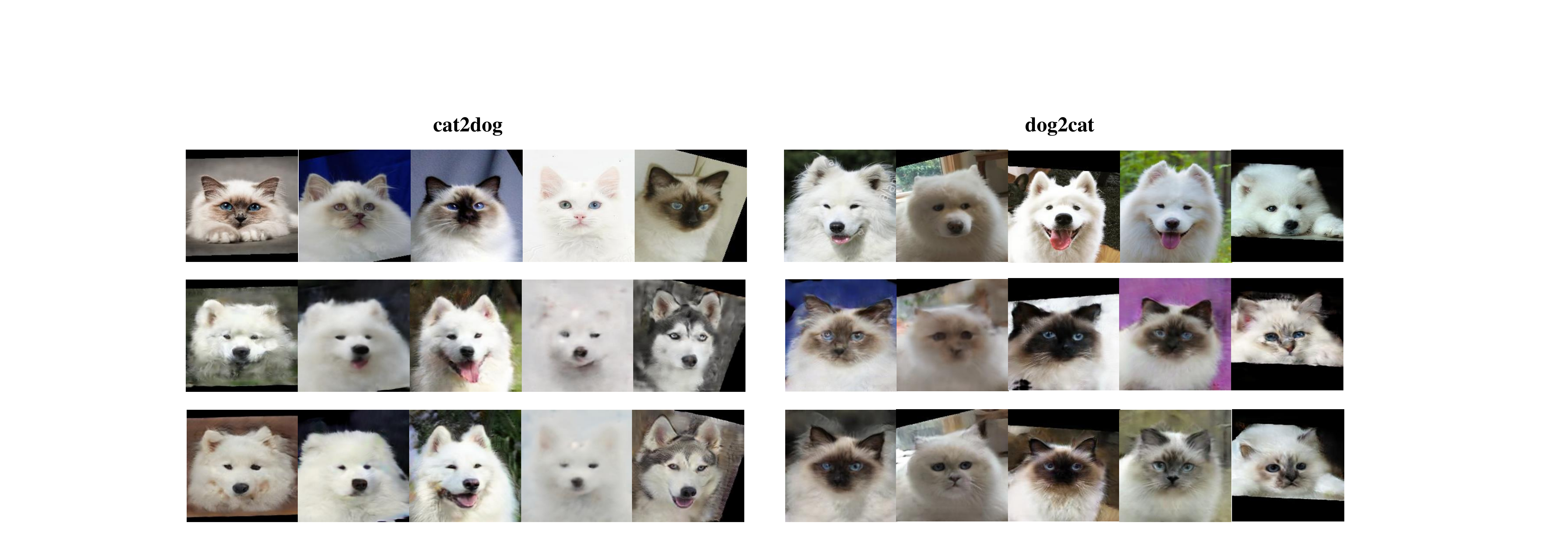}
    \caption{Visual comparisons on cat2dog and dog2cat. \textbf{First row}: input images. \textbf{Second row}: images generated by U-GAT-IT. \textbf{Third row}: images generated by FQ-U-GAT-IT.}
    \label{fig:i2i_cat}
\end{figure*}

\begin{figure*}[!htbp]
    \centering
    \includegraphics[width=0.98\textwidth]{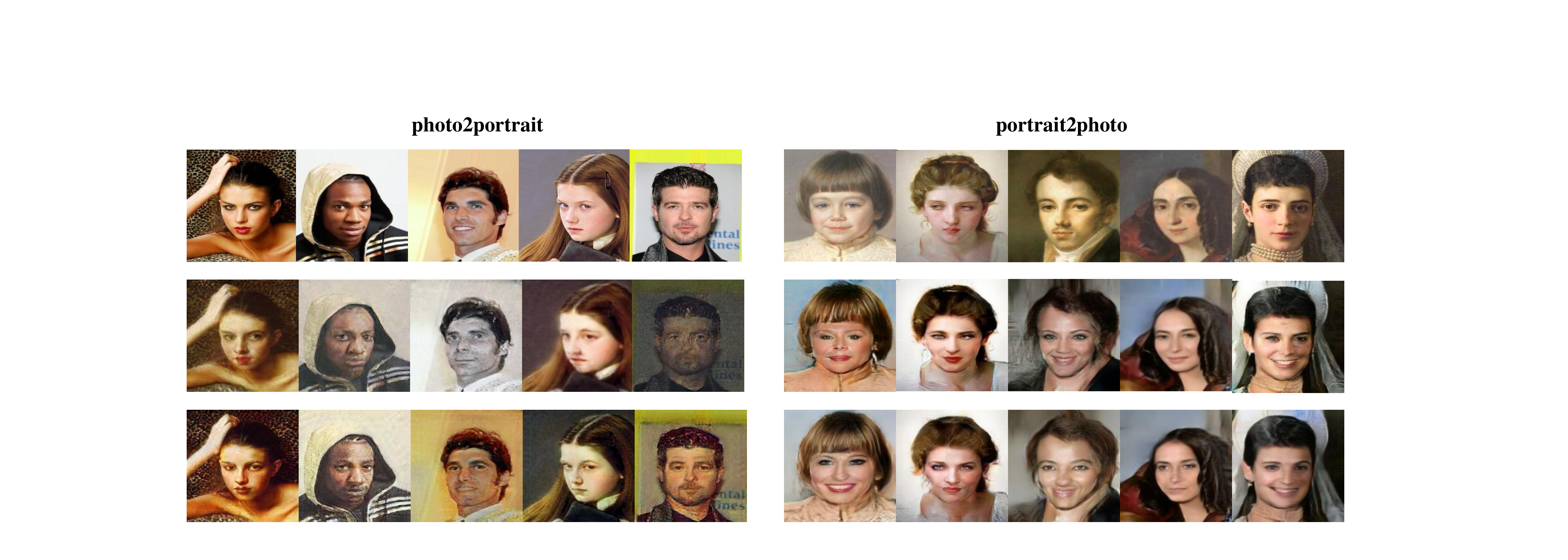}
    \caption{Visual comparisons on photo2portrait and portrait2photo. \textbf{First row}: input images. \textbf{Second row}: images generated by U-GAT-IT. \textbf{Third row}: images generated by FQ-U-GAT-IT.}
    \label{fig:i2i_portrait}
\end{figure*}

\begin{figure*}[!htbp]
    \centering
    \includegraphics[width=0.98\textwidth]{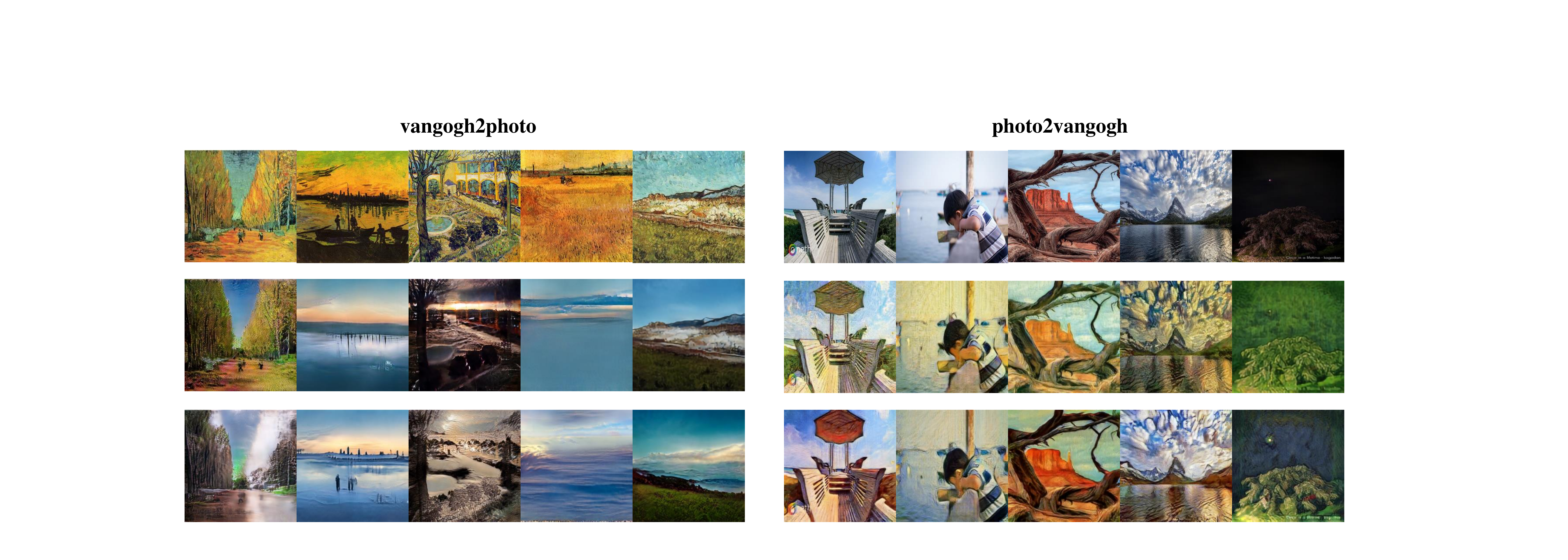}
    \caption{Visual comparisons on vangogh2photo and photo2vangogh. \textbf{First row}: input images. \textbf{Second row}: images generated by U-GAT-IT. \textbf{Third row}: images generated by FQ-U-GAT-IT.}
    \label{fig:i2i_vangogh}
\end{figure*}

\begin{figure*}[!htbp]
    \centering
    \includegraphics[width=0.98\textwidth]{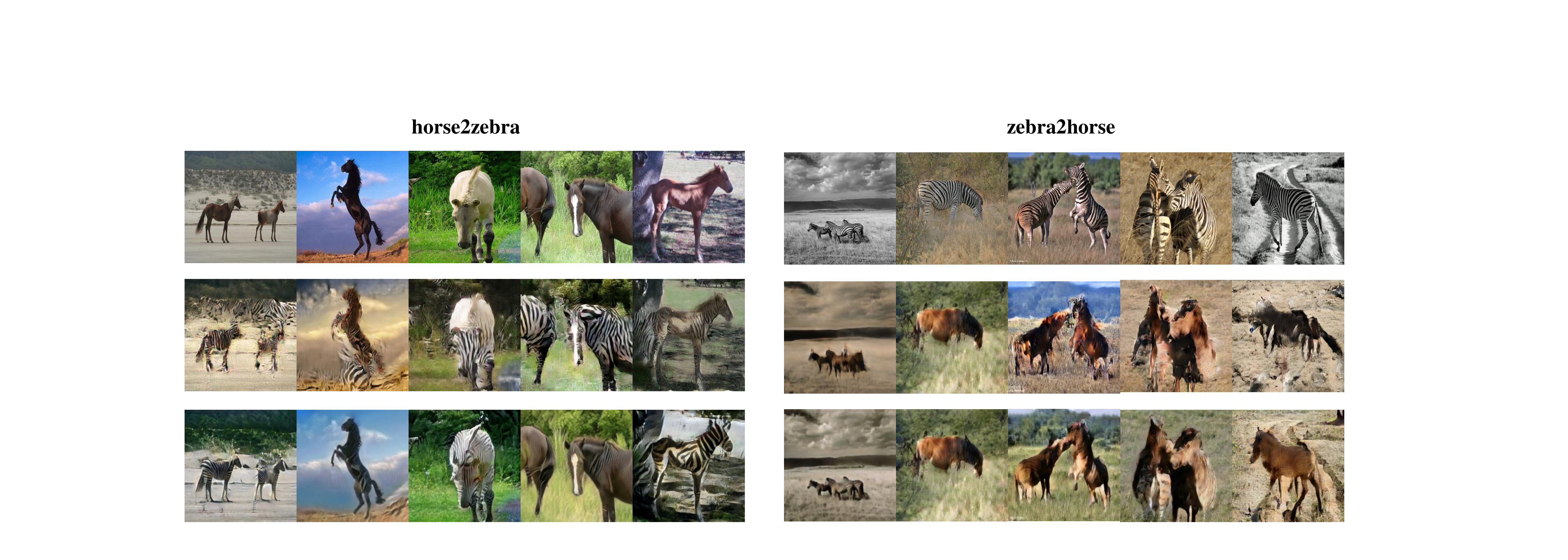}
    \caption{Visual comparisons on horse2zebra and zebra2horse. \textbf{First row}: input images. \textbf{Second row}: images generated by U-GAT-IT. \textbf{Third row}: images generated by FQ-U-GAT-IT. For the horse2zebra translation, U-GAT-IT tends to focus on the texture of zebra but corrupt most details. On contrast, FQ-U-GAT-IT focuses on the horse itself and protect other details. So, FQ-U-GAT-IT fails in some cases (the 4th column) but owns a low KID value.}
    \label{fig:i2i_horse2zebra}
\end{figure*}

We show more translated images: selfie2anime and anime2selfie in Figure~\ref{fig:i2i_anime}, cat2dog and dog2cat in Figure~\ref{fig:i2i_cat}, photo2portrait and portrait2photo in Figure~\ref{fig:i2i_portrait}, vangogh2photo and photo2vangogh in Figure~\ref{fig:i2i_vangogh}, horse2zebra and zebra2horse in Figure~\ref{fig:i2i_horse2zebra}.

\subsection{AMT interface design}
\begin{figure*}[!htbp]
    \centering
    \tcbox{\includegraphics[width=0.98\textwidth]{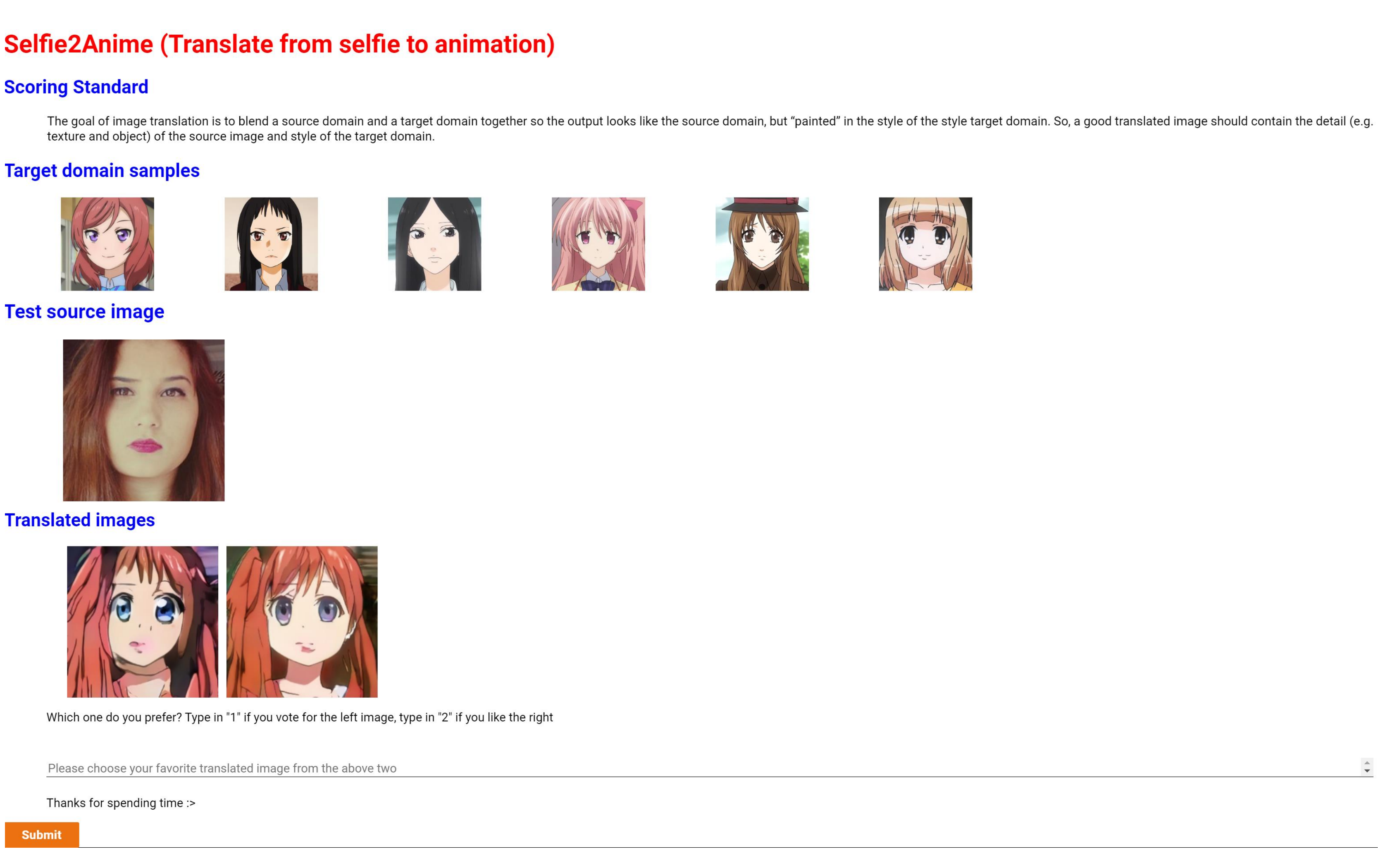}}
    \caption{Interface used for human perceptual study on AMT. }
    \label{fig:amt_interface}
\end{figure*}

The webpage interface used for human evaluation is shown in Figure~\ref{fig:amt_interface}.


\begin{table}[!t]
    \centering
    \caption{BigGAN architecture for $32\times32$ images, $\texttt{ch}=64$. FQ has been added into different ResBlock layers of discriminator.}
    \vspace{0.1in}
    \begin{subtable}{.8\linewidth}
        \centering
        \begin{tabular}{c}
            \toprule
            $z \in \mathbb{R}^{120} \sim \mathcal{N}(0,1)$ \\
            Embed($y$) $\in \mathbb{R}^{128}$ \\\midrule
            Linear $(20 + 128)  \xrightarrow{} 4 \times 4 \times 16ch$\\ \midrule
            ResBlock up $4ch \xrightarrow{} 4ch$ \\ \midrule
            ResBlock up $4ch \xrightarrow{} 4ch$ \\ \midrule
            ResBlock up $4ch \xrightarrow{} 4ch$ \\ \midrule
            BN, ReLU, $3 \times 3$ Conv $ch \xrightarrow{} 3$ \\ \midrule
            Tanh \\ \bottomrule
        \end{tabular}
        \caption{\bf Generator}
    \end{subtable}%
    \vspace{5mm}
    \begin{subtable}{.8\linewidth}
        \centering
        \begin{tabular}{c}
            \toprule
            RGB image $x \in \mathbb{R}^{32\times32\times3}$ \\ \midrule
            Non-Local Block ($64 \times 64$) \\ \midrule
            ResBlock down $4ch \xrightarrow{} 4ch$ \\ \midrule
            ResBlock down $4ch \xrightarrow{} 4ch$ \\ \midrule
            ResBlock $4ch \xrightarrow{} 4ch$ \\ \midrule
            ResBlock $4ch \xrightarrow{} 4ch$ \\ \midrule
            ReLU, Global sum pooling \\ \midrule
            Embed($y$)·$\mathbf{h}$ + (linear $\xrightarrow{}$ 1) \\ \bottomrule
        \end{tabular}
        \caption{\bf Discriminator}
    \end{subtable}%
    \label{tab:arch_cifar}
    \end{table}
\begin{table}[!t]
    \centering
    \caption{Discriminator architecture in StyleGAN and StyleGAN2}
    \vspace{0.1in}
    \begin{tabular}{c|c}
        \toprule
       Blocks-\# & Input $\xrightarrow{}$ Output shape \\\midrule
       1024 & $(1024, 1024, 3) \xrightarrow{Conv} (512, 512, 32)$  \\
       512 & $(512, 512, 32) \xrightarrow{Conv} (256, 256, 64)$  \\
       256 & $(256, 256, 64) \xrightarrow{Conv} (128, 128, 128)$  \\
       128 & $(128, 128, 128) \xrightarrow{Conv} (64, 64, 256)$  \\
       64 & $(64, 64, 256) \xrightarrow{Conv} (32, 32, 512)$  \\
       32 & $(32, 32, 512) \xrightarrow{Conv} (16, 16, 512)$  \\
       16 & $(16, 16, 512) \xrightarrow{Conv} (8, 8, 512)$  \\
       8 & $(8, 8, 512) \xrightarrow{Conv} (4, 4, 512)$  \\
       4 & $(4, 4, 512) \xrightarrow{Conv} (512)$  \\
       Output & $(512) \xrightarrow{Dense} (1)$  \\
    \bottomrule
    \end{tabular}
    \label{tab:stylegan_arch}
    \end{table}
\begin{table}[!htbp]
    \centering
    \caption{BigGAN architecture for $128\times128$ images, $ch=64$.}
    \vspace{0.1in}
    \begin{subtable}{.8\linewidth}
        \centering
        \begin{tabular}{c}
            \toprule
            $z \in \mathbb{R}^{120} \sim \mathcal{N}(0,1)$ \\
            Embed($y$) $\in \mathbb{R}^{128}$ \\\midrule
            Linear $(20 + 128)  \xrightarrow{} 4 \times 4 \times 16ch$\\ \midrule
            ResBlock up $16ch \xrightarrow{} 16ch$ \\ \midrule
            ResBlock up $16ch \xrightarrow{} 8ch$ \\ \midrule
            ResBlock up $8ch \xrightarrow{} 4ch$ \\ \midrule
            ResBlock up $4ch \xrightarrow{} 2ch$ \\ \midrule
            Non-Local Block ($64 \times 64$) \\ \midrule
            ResBlock up $2ch \xrightarrow{} ch$ \\ \midrule
            BN, ReLU, $3 \times 3$ Conv $ch \xrightarrow{} 3$ \\ \midrule
            Tanh \\ \bottomrule
        \end{tabular}
        \caption{\bf Generator}
    \end{subtable}%
    \vspace{5mm}
    \begin{subtable}{.8\linewidth}
        \centering
        \begin{tabular}{c}
            \toprule
            RGB image $x \in \mathbb{R}^{128\times128\times3}$ \\ \midrule
            ResBlock up $ch \xrightarrow{} 2ch$ \\ \midrule
            Non-Local Block ($64 \times 64$) \\ \midrule
            ResBlock down $2ch \xrightarrow{} 4ch$ \\ \midrule
            \textcolor{blue}{$FQ(K=2^{10}, 4ch)$} \\ \midrule
            ResBlock down $4ch \xrightarrow{} 8ch$ \\ \midrule
            ResBlock down $8ch \xrightarrow{} 16ch$ \\ \midrule
            ResBlock down $16ch \xrightarrow{} 16ch$ \\ \midrule
            ResBlock $16ch \xrightarrow{} 16ch$ \\ \midrule
            ReLU, Global sum pooling \\ \midrule
            Embed($y$)·$\mathbf{h}$ + (linear $\xrightarrow{}$ 1) \\ \bottomrule
        \end{tabular}
        \caption{\bf Discriminator}
    \end{subtable}%
    \label{tab:arch_imagenet}
    \end{table}

\begin{table}[!t]
    \centering
    \caption{Modified global discriminator of U-GAT-IT (CAM: Class activation maps \citep{zhou2016learning})}
    \vspace{0.1in}
    \begin{tabular}{c|c}
        \toprule
        Parts & Input $\xrightarrow{}$ Output shape \\\midrule
    \multirow{7}{*}{Encoder Down-sampling}  & $(h, w, 3) \xrightarrow{} (\frac{h}{2}, \frac{w}{2}, 64)$  \\
        &  $ (\frac{h}{2}, \frac{w}{2}, 64) \xrightarrow{}  (\frac{h}{4}, \frac{w}{4}, 128)$  \\ 
        & $(\frac{h}{4}, \frac{w}{4}, 128) \xrightarrow{} (\frac{h}{8}, \frac{w}{8}, 256)$  \\ & \textcolor{blue}{$FQ(K=2^{10}, 256)$} \\
        & $(\frac{h}{8}, \frac{w}{8}, 256) \xrightarrow{} (\frac{h}{16}, \frac{w}{16}, 512)$  \\ 
        & $(\frac{h}{16}, \frac{w}{16}, 512) \xrightarrow{} (\frac{h}{32}, \frac{w}{32}, 1024)$  \\
        & $(\frac{h}{32}, \frac{w}{32}, 1024) \xrightarrow{} (\frac{h}{32}, \frac{w}{32}, 2048)$  \\ \midrule 
        \multirow{2}{*}{CAM of Discriminator} & $(\frac{h}{32}, \frac{w}{32}, 1024) \xrightarrow{} (\frac{h}{32}, \frac{w}{32}, 4096)$ \\ 
        & $(\frac{h}{32}, \frac{w}{32}, 4096) \xrightarrow{} (\frac{h}{32}, \frac{w}{32}, 2048)$ \\ \midrule
       Classifier & $(\frac{h}{32}, \frac{w}{32}, 2048) \xrightarrow{} (\frac{h}{32}, \frac{w}{32}, 1)$
        \\ \bottomrule
    \end{tabular}

    \label{tab:arch_ugatit}
    \end{table}

\end{document}